\documentclass[a4paper,12pt]{article}
\usepackage[utf8]{inputenc}
\usepackage[T1]{fontenc}
\usepackage{graphicx}
\usepackage{hyperref}
\usepackage{geometry}
\usepackage{amsmath}
\usepackage{ulem}

\title{Three-Factor Learning in Spiking Neural Networks: An Overview of Methods and Trends from a Machine Learning Perspective}
\author{
    Szymon Mazurek$^{1,3,4}$, 
    Jakub Caputa$^{1}$, 
    Jan K. Argasiński$^{2,4}$, 
    Maciej Wielgosz$^{1,}$\thanks{Corresponding author: wielgosz@agh.edu.pl}
}

\begin{document}
\date{}
\maketitle


\begin{center}
$^{1}$ AGH University of Krakow, Poland\\
Email: \{szmazurek, jakubcaputa, wielgosz\}@agh.edu.pl\\[5pt]
$^{2}$ Jagiellonian University, Poland\\
Email: jan.argasinski@uj.edu.pl \\
[5pt]
$^{3}$ ACC Cyfronet AGH, Poland \\
Email: s.mazurek@cyfronet.pl \\
[5pt]
$^{4}$ Sano - Centre for Computational Medicine, Poland \\
Email: j.argasinski@sanoscience.org
\end{center}
\begin{abstract}
Three-factor learning rules in Spiking Neural Networks (SNNs) have emerged as a crucial extension to traditional Hebbian learning and Spike-Timing-Dependent Plasticity (STDP), incorporating neuromodulatory signals to improve adaptation and learning efficiency. These mechanisms enhance biological plausibility and facilitate improved credit assignment in artificial neural systems. This paper \uline{takes a view on this topic from a machine learning perspective}, providing an overview of recent advances in three-factor learning, discusses theoretical foundations, algorithmic implementations, and their relevance to reinforcement learning and neuromorphic computing. In addition, we explore interdisciplinary approaches, scalability challenges, and potential applications in robotics, cognitive modeling, and AI systems. Finally, we highlight key research gaps and propose future directions for bridging the gap between neuroscience and artificial intelligence.

\end{abstract}

\section{Introduction}

In recent years, \textbf{Spiking Neural Networks (SNNs)} have emerged as a promising paradigm in artificial intelligence, inspired by the way biological neurons communicate through discrete spikes \cite{Richards2019, Zenke2015}. Unlike traditional artificial neural networks, SNNs operate in a time-sensitive manner, allowing them to model temporal patterns and perform energy-efficient computation \cite{Najarro2020, Mozafari2019}. Despite these advantages, many challenges remain in the development of efficient learning rules for SNNs, particularly in scaling these models to complex real-world applications \cite{Bellec2019, Sutton2021} and developing learning algorithms that fully exploit their properties \cite{Yi2023}.

The fundamental biologically inspired learning rule is \textbf{Spike-Timing-Dependent Plasticity (STDP)}\cite{Hebb1949}, where synaptic weights are modified based on the temporal coincidence of pre-synaptic and post-synaptic spikes arriving at the neuron.
An extension of STDP and the main topic of this review is the \textbf{three-factor learning rule}, which incorporates an additional modulatory signal, often representing neuromodulators such as dopamine \cite{Fremaux2016, Gerstner2018_2, Pawlak2010}. This third factor plays a crucial role in guiding plasticity by integrating global contextual information, allowing the network to learn both reward signals and environmental feedback \cite{Tiesinga2001, Florian2005}.

\begin{figure}[h]
    \centering
    \includegraphics[width=0.5\linewidth]{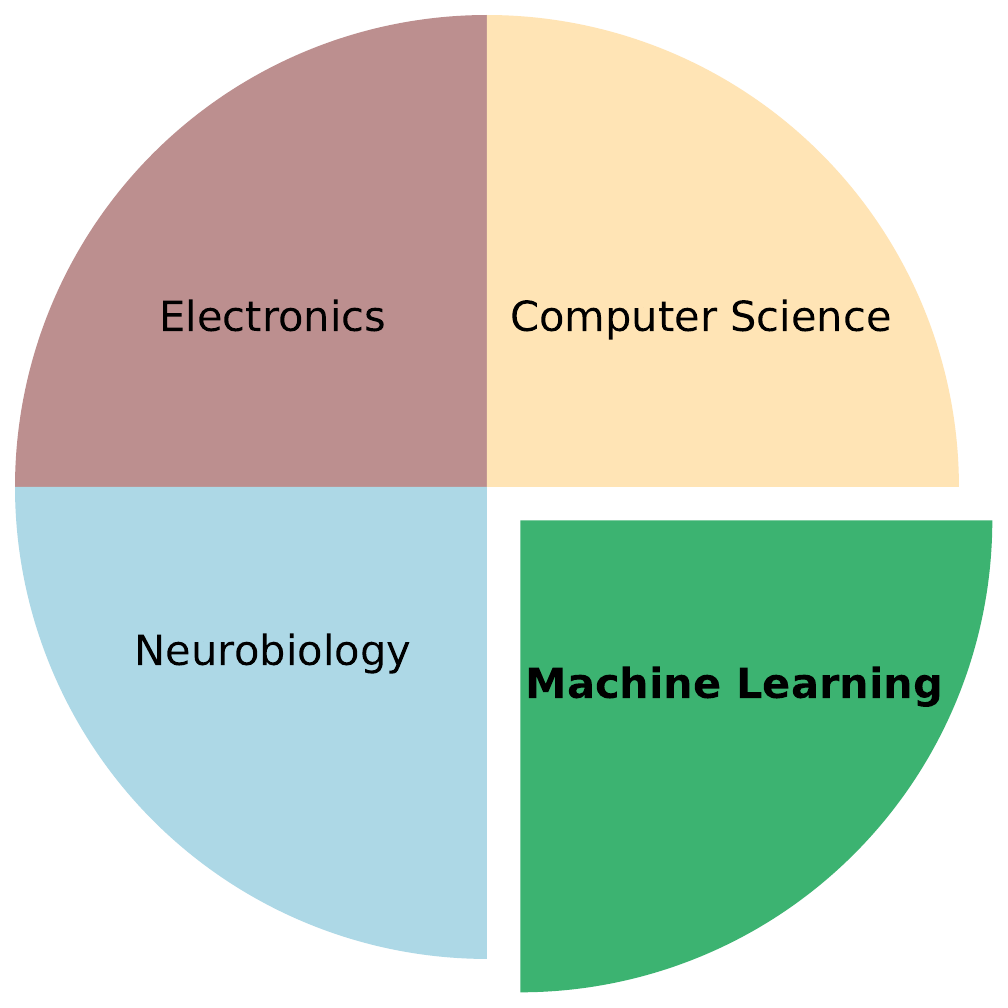}
    \caption{Spiking Neural Network domain and its key components. In our work, we focus explicitly on the highlighted subdomain - machine learning.}
    \label{fig:4-pie-plot}
\end{figure}

SNN research is a complex domain that operates at the intersection of electronics, computer science, neurobiology, and machine learning, as illustrated in Fig.\ref{fig:4-pie-plot}. Each of these fields contributes essential foundational principles for SNN development: electronics enables the design of neuromorphic hardware, computer science provides algorithmic frameworks, neurobiology offers biologically inspired learning mechanisms, and machine learning drives optimization and scalability. However, integrating these diverse disciplines into a cohesive framework remains a significant challenge. Differences in methodologies, terminology, and research priorities often create gaps between theoretical neuroscience, computational modeling, and hardware implementations. Machine learning, while central to modern advancements, must balance biological plausibility, computational efficiency, and hardware feasibility to create SNN models that are both functionally powerful and practically deployable. Fig. \ref{fig:4-pie-plot} visually represents this complexity, highlighting that while machine learning is a dominant force, it cannot operate in isolation from the other three fields.

Consequently, this paper aims to provide a review of the state of three-factor learning from a machine learning perspective. In parallel, we want to emphasize the complexity of the discipline and the necessary cross-domain research collaboration to properly merge knowledge from all fields shown in Fig.\ref{fig:4-pie-plot}. To achieve this, a series of papers were analyzed,
highlighting the following:

\begin{itemize} 
    \item \textbf{Interdisciplinary Perspectives:} This review explores three-factor learning in spiking neural networks (SNNs) from both theoretical and practical viewpoints, highlighting the convergence of neuroscience and artificial intelligence \cite{Fremaux2013, Pedrosa2017}.
    \item \textbf{Neuromodulatory Mechanisms:} It emphasizes how the third factor, global modulatory signals such as dopamine, can steer synaptic changes beyond standard STDP, improving adaptive behaviors and learning efficacy \cite{Brzosko2019, Aljadeff2019}.
    \item \textbf{Cognitive Modeling and Robotics:} The discussion covers real-world implications, showcasing how three-factor learning enables robust, context-aware SNN applications in tasks ranging from decision making to autonomous navigation \cite{Sporns2002, Alnajjar2009}.
    \item \textbf{Scalability and Encoding Strategies:} It addresses key challenges in scaling three-factor learning to larger networks, including computational constraints, diverse spike-encoding approaches, and the need for efficient hardware support \cite{Hoerzer2012, Kopsick2022}.
    \item \textbf{Future Opportunities:} Finally, it describes promising avenues for cross-domain research, bridging the gaps between theoretical models and applied technologies to further advance three-factor learning in SNN \cite{Schmidgall2023_2, Gerstner2018}.
\end{itemize}

In addition to summarizing current advances, this paper identifies research gaps, offering recommendations for future directions that address these limitations through interdisciplinary synthesis. Due to the fact that three-factor learning and SNNs are still a widely unexplored and emerging discipline, our review approach does not strictly conform to any systematic review methodologies, such as the PRISMA or Kitchenham guidelines \cite{Page20220Prisma,Kitchenham2009Systematic}. In the following chapters, we present and evaluate various categories that highlight different aspects of the reviewed papers. The categorization we propose is inherently approximate, as the boundaries between categories are often indistinct and overlap. Nevertheless, this classification represents our best effort to introduce a structured framework for reasoning about this highly heterogeneous, rapidly evolving, and still-nascent field.

\section{Theoretical Foundations of Three-Factor Learning}

\begin{figure}
    \centering
    \includegraphics[width=1\linewidth]{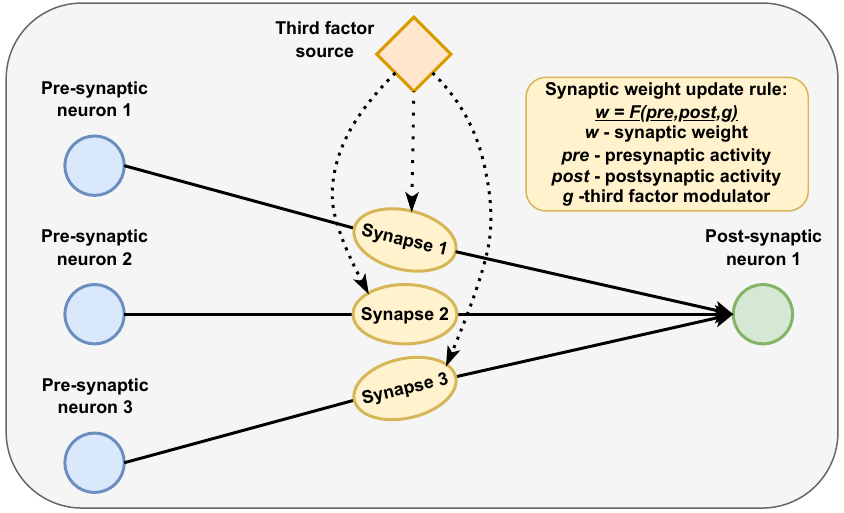}
    \caption{General overview of three-factor learning principle: synaptic weights are modified based on local activity and influence of the third factor. }
    \label{fig:three-factor-learning-principle}
\end{figure}

The concept of three-factor learning is presented in Fig.\ref{fig:three-factor-learning-principle}, with its roots appealing to biological neural mechanisms. In the human brain, learning is not solely driven by local synaptic activity but is heavily influenced by global signals, such as neuromodulators: dopamine, serotonin, and acetylcholine \cite{Marder2002, Brzosko2019}. These signals regulate synaptic plasticity, allowing the brain to adapt based on rewards, motivation, and contextual information \cite{Lavigne2008, Pawlak2010}. A visualization of this rule is shown in Fig.\ref{fig:three-factor-learning-principle}.

Research in neuroscience has shown that STDP is insufficient to fully explain complex learning behaviors \cite{Suvrathan2018, Fremaux2016}. The introduction of a third factor in learning models aligns with the findings on how global neuromodulatory systems interact with local synaptic processes. For example, dopamine has been linked to reward-based learning, playing a critical role in the reinforcement learning mechanisms observed in biological systems \cite{Fremaux2010, Soltani2006}. This inspiration has led to the development of computational models that attempt to replicate these dynamics in artificial SNNs \cite{Talanov2015, Florian2005}.

The concept of three-factor learning has evolved over several decades and is shaped by both theoretical and experimental advances. Early research on synaptic plasticity emphasized two-factor models, such as Hebbian Learning and Spike-Timing Dependent Plasticity (STDP) \cite{Hebb1949, Tiesinga2001}. However, these models faced limitations in explaining reward-driven behaviors and long-term adaptations observed in biological systems \cite{Froemke2015, Brzosko2019}.

In the late twentieth century, neuroscientists began to uncover the role of neuromodulators such as dopamine in reinforcement learning \cite{Gruber2003, Edeline2012}. The discovery of dopamine involvement in reward prediction errors led to the formulation of models that incorporated an additional global factor in synaptic updates \cite{Fremaux2013, Richards2018}. These models demonstrated improved learning capabilities in both simulations and empirical studies \cite{Foncelle2018, Sutton2021}.

By the early 2000s, the computational neuroscience and machine learning communities started to converge on the importance of three-factor learning. Research focused on developing algorithms that balance local synaptic updates with global feedback signals, resulting in enhanced performance for tasks that require long-term planning, decision making, and contextual adaptation \cite{Legenstein2009, Hoerzer2012}. Today, three-factor learning is recognized as an important component in bridging the gap between biological plausibility and artificial learning systems \cite{Gerstner2018_2, Mozafari2019_2}.

Currently, the field of three-factor learning in Spiking Neural Networks is characterized by a growing consensus on the importance of integrating local and global learning signals \cite{Whittington2019, Pedrosa2017}. Researchers have developed various algorithms that take advantage of neuromodulatory influences to improve network adaptability and learning efficiency \cite{Yuan2019, Najarro2020}. These advancements have contributed to improved performance in tasks requiring temporal memory, reward-based learning, and complex decision-making \cite{Bellec2019, Liu2021}.

Experimental studies have shown that incorporating three-factor learning mechanisms can enhance the stability of network dynamics \cite{Durstewitz, Aljadeff2019}. This is particularly important in tasks where networks must balance exploration and exploitation or operate under delayed reward conditions \cite{Parussel2007, Sporns2002}. Computational models now commonly simulate neuromodulatory effects, enabling more biologically plausible learning processes \cite{Hasselmol2018, Nolan2010}.

Despite these improvements, several challenges remain. Scalability for large networks, the design of efficient hardware platforms and the use of real-world datasets are areas where further research is needed \cite{Schmidgall2023, Vigneron2020}. Furthermore, there is ongoing work to unify disparate approaches under a cohesive theoretical framework that connects biological mechanisms with artificial implementations \cite{Barry2022, Allred2020}. As a result, current research is focused on cross-disciplinary efforts that aim to refine both the theoretical understanding and practical applications of three-factor learning \cite{Gerstner2018_2, Richards2019}.

\subsection{Neuromodulatory Influence on Synaptic Plasticity}

Neuromodulation plays a crucial role in synaptic plasticity by integrating intrinsic and extrinsic signals that affect neuronal interactions and learning dynamics. In this section, we will formalize the learning rules discussed and show how synapse modulation could be manifested on the basis of insights from neuroscience.

\paragraph{Spike-Time Dependent Plasticity and effects of third factor modulation}

The Spike-Time Dependent Plasticity (STDP) can be regarded as application of Hebb's postulate \cite{Hebb1949}, worded as "neurons that fire together, wire together". This intuitive statement indicates that synapses for which presynaptic and postsynaptic spiking activity coincides temporally result in a synaptic weight change:

\begin{equation}
    \Delta w_t = 
        \begin{cases} 
             e^{-\Delta t / \tau_+}, & \Delta t > 0 \\
             e^{\Delta t / \tau_-}, & \Delta t < 0
        \end{cases}
\end{equation}

where:
\begin{itemize}
    \item \( \Delta w_t \) is the change in synaptic weight at time $t$,
    \item \( \Delta t = t_{\text{post}} - t_{\text{pre}} \) is the timing difference between pre- and postsynaptic spikes,
    \item \( \tau_+ \) and \( \tau_- \) are time constants that control the decay of the STDP window.
\end{itemize}

Thus, if pre-synaptic spikes directly precede postsynaptic spikes, we observe long-term potentiation (LTP), resulting in an increase in synaptic weight. If the opposite is true, long-term depression (LTD) occurs and synaptic weight decreases. The shape of LTD and LTP windows is controlled by the hyperparameter $\tau$.

\begin{figure}
    \centering
    \includegraphics[width=1\linewidth]{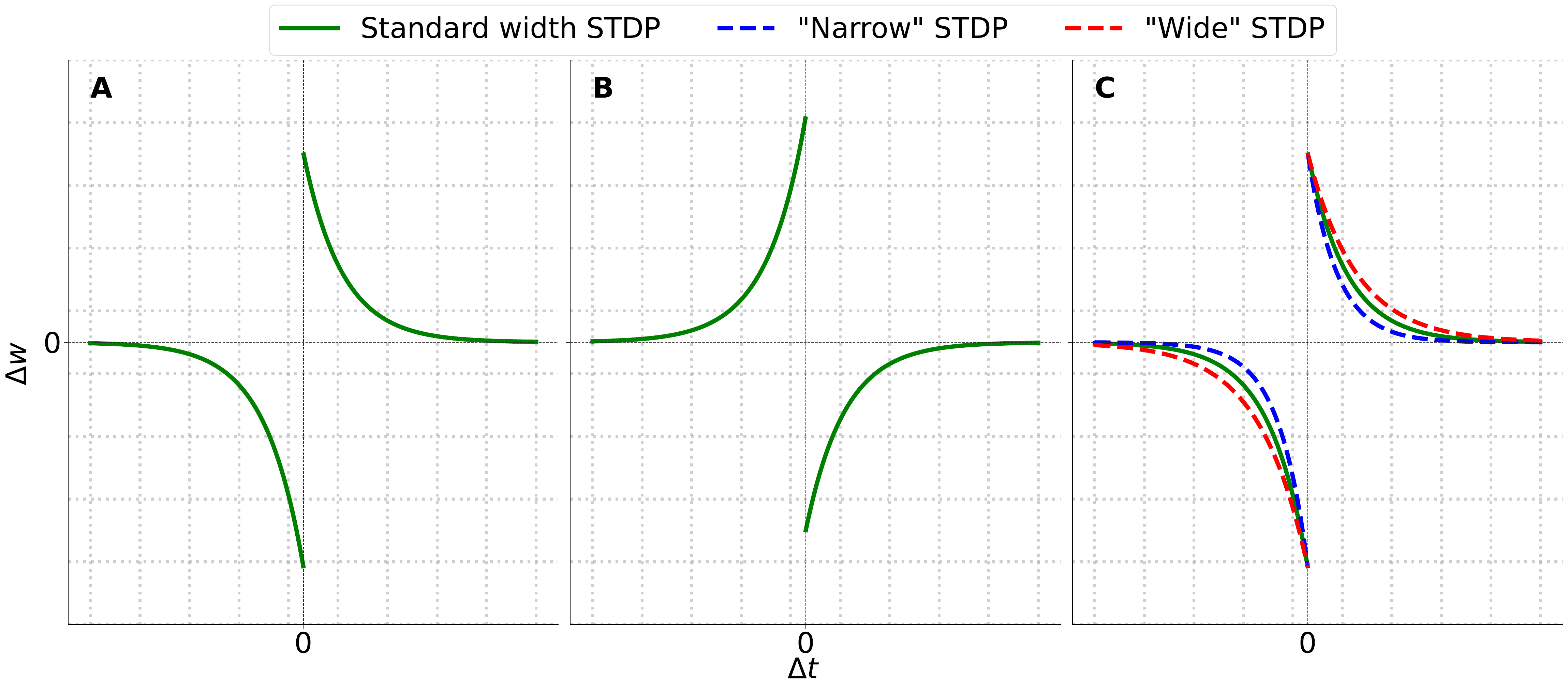}
    \caption{Possible influences of third factor on Spike-Timing Dependent Plasticity (STDP) learning rule. The \textbf{A} plot shows baseline STDP, where synaptic weight change (\(\Delta w\)) depends on the relative timing (\(\Delta t\)) of pre- and post-synaptic spikes. The \textbf{B} plot illustrates reversed STDP, where the LTD and LTP polarities are flipped. The \textbf{C} plot demonstrates STDP shape modulation, where neuromodulatory factors influence the temporal profile of plasticity, modifying the learning window width.}
    \label{fig:stdp_modulated}
\end{figure}

In general, we can simplify the above equation as a function of presynaptic and postsynaptic spikes:

\begin{equation}
    \Delta w_t = H(t_{pre}, t_{post}),
\end{equation}
where \( H(\cdot) \) is a function governing synaptic plasticity. The introduction of the third factor extends the STDP rule in the following form:

\begin{equation}
    \Delta w_t = H(t_{pre}, t_{post}, g_t),
\end{equation}
where $g$ is the modulatory signal affecting the neuron at time $t$. This third factor signal can broadly influence the dynamics of base synaptic plasticity \cite{Marder2002}. Based on neurobiological knowledge \cite{Pawlak2010}, synaptic neuromodulation can induce effects such as amplification of the weight change; reversal of the STPD window (swapping LTD with LTP on the timescale); changing the width of the LTP and LTD windows, or even gate the occurrence of a synaptic weight modification. Visualization of these exemplary effects can be seen in Fig. \ref{fig:stdp_modulated}.

\subsection{Spatial and temporal aspects of third factor modulation}
\begin{figure}[h]
    \centering
    \includegraphics[width=0.8\linewidth]{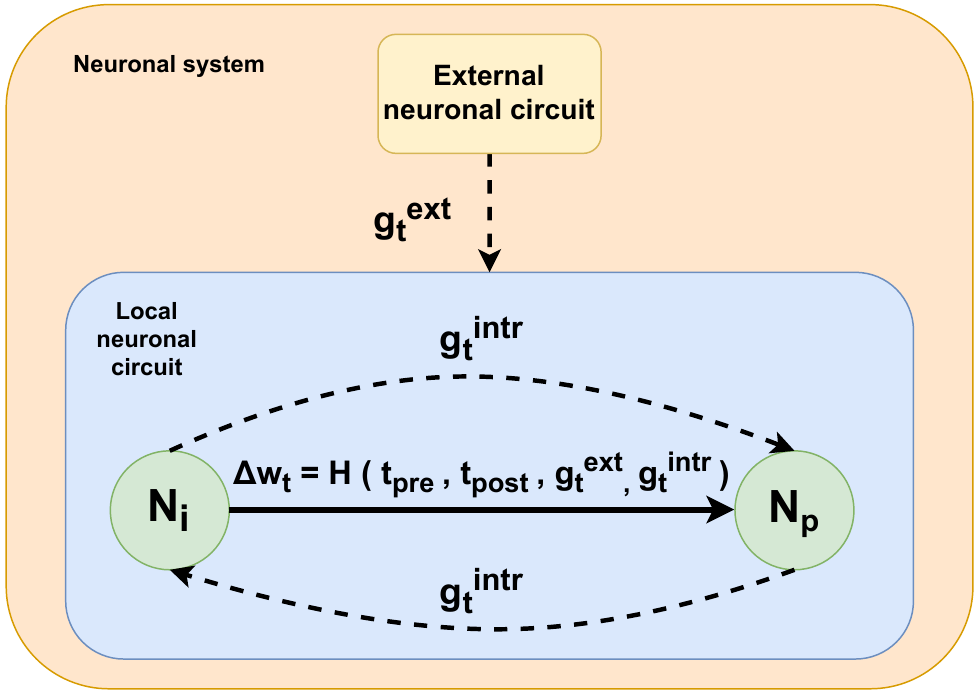}
    \caption{Different sources of the top-level third factor: the signal can be emitted intrinsically between neurons in the same neuronal circuit or extrinsically, when the signal arrives from outside of the circuit. Neurons that are coinciding either spatially, functionally or morphologically are considered to be in the same circuit \cite{Marder2002}.}
    \label{fig:3factor-extrinsic-intrinsic}
\end{figure}

One of the fundamental problems with respect to the use of three-factor learning rules is the spatial and temporal aspects of modulatory signal effectiveness. The relationships between neuromodulators in these domains are notoriously complex and difficult to observe in biological systems \cite{Marder2002}.
Although the temporal properties of modulatory signals have already been incorporated into the discussed equations, spatial properties should also be included. Thus, we refer to the concepts of intrinsic and extrinsic neuromodulation, graphically described in Fig. \ref{fig:3factor-extrinsic-intrinsic}.

\paragraph{Intrinsic Neuromodulation}

Intrinsic neuromodulatory signals are exchanged between neurons within the same neuronal circuit. Neurons are considered to be within the same local network if they are coinciding either spatially, functionally, or morphologically \cite{Marder2002}. It can be said that they serve as a "memory" that adjusts the dynamics of the local network based on its recent and current activity \cite{Lizbinski2018}. Intrinsic neuromodulatory signal can be described as:

\begin{equation}
    g_t^{\text{intr}} = f_{\text{intr}} \left( N^i_t, N^p_{t}, S^{\text{intr}}_t \right),
\end{equation}
where \( f_{\text{intr}} \) describes how the state $S^{\text{intr}}$ of the internal network determines the local neuromodulatory effect between neurons $N^i$ and $N^p$ at time $t$.

\paragraph{Extrinsic Neuromodulation}

Signals from external neural networks influence other circuits by providing information on their ongoing activity \cite{Lizbinski2018}. The modulatory signal that affects a population of neurons $P$ at time $t$ can be defined as:

\begin{equation}
    g_t^{\text{ext}} = f_{\text{ext}} \left( P_t, S_t^{\text{ext}} \right),
\end{equation}
where $S^{\text{ext}}$ is a state of external neuronal circuit.

\paragraph{Extended Synaptic Plasticity Function}

Based on the concepts discussed so far, we can formulate a more detailed model for synaptic plasticity, which incorporates weighted contributions from different factors, as well as spatial and temporal properties of all signals influencing plasticity of a given synapse.

\begin{equation}
    \Delta w_t = H(t_{pre}, t_{post}, g_t^{intr},g_t^{ext})
\end{equation}



This formulation highlights how intrinsic and extrinsic neuromodulatory factors contribute to synaptic plasticity, ultimately shaping learning and adaptive behaviors in neural networks. We note that the presented equations can be further extended, and their presented derivation is not exhaustive due to complexity of the plasticy phenomena.

\subsection{A note on backpropagation through time}

While three-factor learning rules offer a biologically plausible approach to training SNNs, it is important to acknowledge the role of Backpropagation Through Time (BPTT) based on surrogate gradients \cite{Werbos1990BPTT, Neftci2019Surrogate}. Both of those topics are extensive and beyond the scope of this article, yet we will briefly describe them to provide an overview of the problems encountered when training SNNs and their relationship with biologically plausible learning methods.

The surrogate gradient method enables gradient-based learning in SNNs by approximating neuron spiking activity with continuous function, which allows error backpropagation \cite{Rumelhart1986backprop}. BPTT enables one to perform backpropagation in the temporal domain, which is inherent in SNNs. The combination of these methods allows for effective training of deep SNN architectures with the well-known approaches established in deep learning research. Furthermore, the performance achieved when using them is robust, closely compared to that observed with classic deep neural networks \cite{Zenke2015}.

However, BPTT in its standard form faces several challenges. 
Firstly, the computational cost and memory footprint of BPTT can be substantial, especially for long input sequences, due to the need to store neuron states at each time step. This also implies relatively slow processing, as the system is gradient computation for consecutive steps is sequential in nature.
Secondly, BPTT can suffer from problems with the stability of the training process, as vanishing or exploding gradients can hinder the learning of long-range temporal dependencies \cite{Pascanu2013Vanishing}.

In general, BPTT is considered biologically implausible, as it deviates from the local learning mechanisms observed in the brain \cite{Lillicrap2019bpttBrain}. The ongoing research aims to address these limitations by exploring memory-efficient BPTT techniques, such as activation checkpointing and truncated BPTT \cite{Beaumont2024checkpointing,Lillicrap2019bpttBrain}, and developing more biologically inspired approximations such as local learning rules and eligibility trace propagation \cite{Bellec2019,Shrestha2019approximating}.

Consequently, the development of alternative learning paradigms, such as three-factor learning rules, is of paramount interest in the field of SNNs. Research on robust and backpropagation-free methods offers the potential to increase speed, scalability, and energy efficiency while maintaining competitive performance. Such advancements would unlock numerous applications for SNNs, particularly in resource-constrained environments and online learning scenarios.

\section{Research Trends}

\subsection{Learning Algorithms}

A wide range of learning algorithms has been explored in SNNs, many inspired by biological mechanisms. Three-factor learning rules have gained prominence, introducing a third element such as neuromodulators or error signals to improve synaptic updates \cite{Gerstner2018, Fremaux2016, Kuśmiesz2017}.  The fundamental motivation behind three-factor learning stems from the need to model biological neural plasticity more accurately. Traditional learning approaches often struggled to capture the complex mechanisms of synaptic modification observed in biological systems. By introducing a third factor, typically a neuromodulatory signal, error signal, or reward signal, researchers have developed more sophisticated learning algorithms that can adapt more dynamically to environmental contexts. In this section, we describe a selection of research articles that demonstrate the applicability of three-factor learning in solving machine learning tasks.

Several key approaches have emerged in the development of three-factor learning rules. In \cite{Fremaux2010}, authors analyze the functional requirements for R-STDP (reward modulated STDP) using a simple set of SRM neurons. They compare R-STDP and R-max STDP, where the reward signal acts as an additional multiplier of the change in synaptic weight, delivered at the end of a task to indicate success. Through trajectory learning and spike train response tasks, they explore the theoretical underpinnings of reward-modulated plasticity, concluding that effective reward-based learning requires a small unsupervised term influence, sensitivity to reward timing, and a reward prediction mechanism.

Reinforcement learning principles have been particularly influential in three-factor learning strategies. In \cite{Fremaux2013}, authors propose a continuous-time actor-critic framework for reinforcement learning in spiking neural networks. They explicitly model temporal credit assignment using temporal difference (TD) learning, where synaptic plasticity is modulated by TD error. The approach integrates value and policy networks with reward-modulated STDP. They evaluated their method in challenging reinforcement learning tasks including Morris water-maze navigation, acrobot, and cartpole simulations, demonstrating the effectiveness of their approach in complex motor control scenarios.

Vasilaki et al. \cite{Vasilaki2009} explored spike-based reinforcement learning in continuous state and action spaces, addressing cases where traditional policy gradient methods fail. They propose a feedforward SNN model in which reward modulates the probability of firing sequences propagating from place cells (representing agent position) to action cells (controlling movement). Synaptic changes are driven by STDP, modulated by a biologically plausible third-factor reward signal. The model is tested in a simulated water maze task, showcasing its potential for sophisticated spatial navigation learning.

More recent developments have pushed the boundaries of three-factor learning. In \cite{Bellec2019}, the authors introduce the eprop algorithm, with the aim of approximate backpropagation through time for global error projection. They take inspiration from the ERN signal preceding error in the brain. The algorithm achieves results close to BPTT on a variety of tasks including speech recognition, word prediction, one-shot learning, and pattern generation.

Building on these foundations, \cite{Liu2021} proposes MDGL, an innovative algorithm to propagate top-down error signals to specific neurons in the network, which then propagate them further in their local neighborhood. Through comprehensive evaluation, they demonstrate that their method closely matches BPTT and outperforms eprop in online learning and pattern generation tasks.

In a notable contribution, \cite{Schmidgall2023} shows an interesting approach of using signal obtained with meta-learning to modulate STDP, which is optimized by gradient descent. The synaptic change occurs when a neuromodulatory signal appears. They demonstrate the robustness of their solution by evaluating the network in T-maze navigation, character recognition, and cue association tasks.

Barry et al. \cite{Barry2022} developed a method using modulated STDP to gate plasticity, introducing a surprise signal derived from error. Their approach involves inducing synaptic changes whenever a surprise signal arrives. Through rigorous testing in continual learning and rule-switching scenarios, they showcase the system's ability to rapidly adapt while maintaining operational stability.

Quintana et al. \cite{Quintana2024ETLP} propose a novel event-based three-factor local plasticity (ETLP) method tailored for online learning with neuromorphic hardware. Their approach features a unique architecture where hidden layers update weights through random matrices, and the output neurons are connected one-to-one to excitatory and inhibitory synapses. Evaluated on pattern recognition tasks using N-MNIST and SHD datasets, ETLP achieves competitive classification accuracy with lower computational complexity compared to global methods like BPTT and eprop.

These algorithms collectively aim to improve performance on tasks that require temporal memory, decision-making, and context-sensitive learning, with a strong emphasis on mimicking the mechanisms found in natural neural systems. This emphasis reflects a broader trend towards integrating both local synaptic updates and global modulatory signals to improve the scalability and efficiency of learning in SNNs \cite{Legenstein2009, Bellec2019}.


\subsection{Datasets}

\begin{figure}[h]
    \centering
    \includegraphics[width=0.8\textwidth]{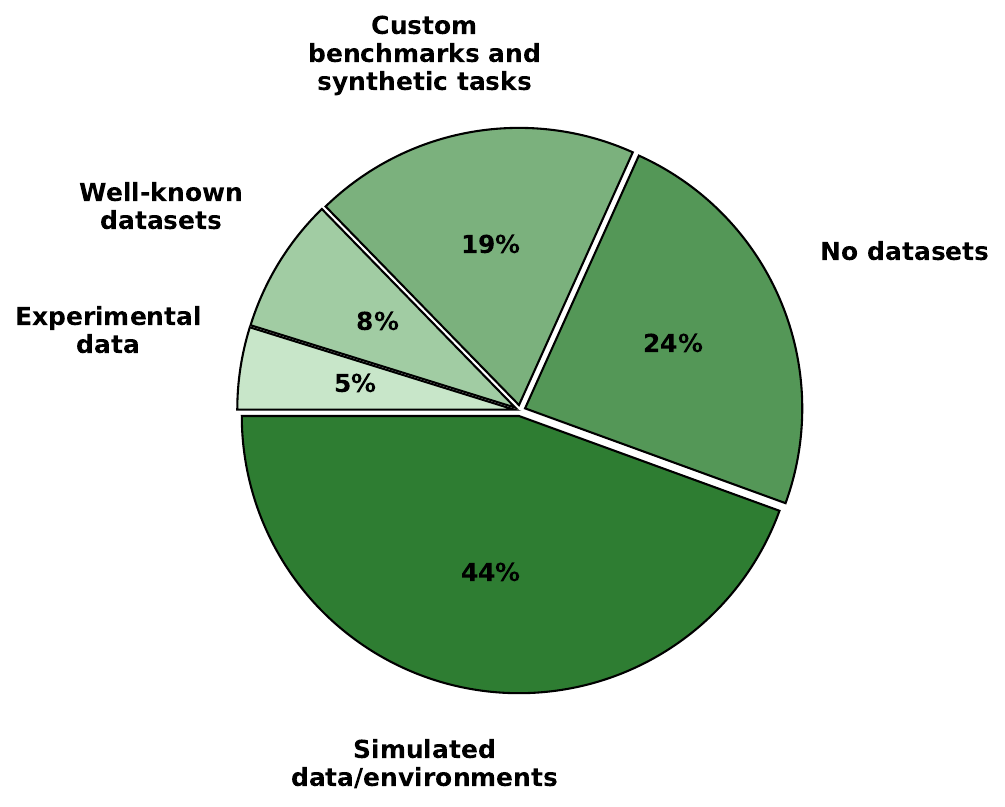}
    \caption{Datasets used in research papers investigating three-factor learning in SNNs.}
    \label{fig:datasets}
\end{figure}

Given the inherent heterogeneity and the involvement of multiple domains in three-factor research, the characteristics and types of datasets used exhibit significant variation, as presented in Fig. \ref{fig:datasets}. A significant number of studies are based on synthetic or custom-designed benchmarks, which allow precise control of experimental variables \cite{Barry2022, Zambrano2021}. These datasets simulate tasks such as navigation (e.g., 1D and 2D multi-target tasks), robotics (e.g., robotic arm reaching and terrain crossing), cognitive tasks (e.g., working memory, decision-making, and attention), and pattern recognition \cite{Schmidgall2023, Hasselmol2018}. Custom tasks like rule-switching, memory-guided saccades, and associative learning are also common \cite{Gruber2003, Avery2014}.

In contrast, some studies \cite{Mozafari2019, Mozafari2019_2, Quintana2024ETLP} incorporate well-known machine learning datasets such as MNIST \cite{Deng2012mnist}, Caltech-256\cite{griffin2022caltech}, ETH-80 \cite{leibe2003analyzing}, and NORB \cite{Lecun2004norb} for image recognition or SHD \cite{Cramer2022SHD}for speech analysis. Some of those datasets, especially in image recognition, are adapted to have the properties of recordings gathered with native neuromorphic sensors. An example of such adoption is the N-MNIST dataset \cite{Orchard2015Nmnist}, used by Quintana et al. \cite{Quintana2024ETLP}.
There are also cases, especially in neuroscience research, in which experimental data from biological studies are used, including cortical slice recordings, optogenetic experiments, and natural scene videos \cite{Edeline2012, Aljadeff2019}. However, such studies focus on discovering biological mechanisms in living systems. Although they remain crucial for the discovery of bioplausible learning mechanisms, such studies often do not attempt directly to train SNNs to solve a specific task using the discovered phenomena.

Thus, it can be observed that real-world datasets remain underutilized, particularly in studies focusing on theoretical models and neural mechanisms \cite{Kopsick2022, Fink2011}. This indicates a trend towards task-specific simulations over standardized benchmarks. In the future, there is a growing need to validate models through greater use of real-world data, ensuring that the proposed learning algorithms are robust in diverse environments and applications \cite{Sutton2021, Sporns2002}. Furthermore, applied SNN research would greatly benefit from establishing a wide set of standard neuromorphic datasets, comparable to MNIST or ImageNet for classic deep learning. The number of such datasets is growing \cite{Cramer2022SHD, Orchard2015Nmnist}, yet it remains a challenge, often demanding specialized neuromorphic hardware and precise experimental setups. In addition, SNN applications span various domains, making standardization a persistent challenge.

\subsection{Application Domains}

Research on three-factor learning in SNNs spans multiple scientific and technological domains, with neuroscience and neurobiology forming the foundation of many studies \cite{Edeline2012, Marder2002, Szatm2010, Soltani2006}. An important line of study is the modeling of neural circuits in brain regions such as the hippocampus, cortex, and basal ganglia, to investigate the mechanisms underlying synaptic plasticity: STDP, neuromodulation, and neurotransmitter influences \cite{Murphy-Royal2023, Durstewitz, Brzosko2019_2, Pedrosa2017}. Biological studies on three-factor learning rules highlight the ability of these algorithms to better capture cognitive processes such as memory, attention, decision-making, and brain rhythms \cite{Gruber2003, Toporikova2011, Aljadeff2019, Gerstner2018_2}. Many of these approaches are validated with experimental data from in vitro and in vivo studies, showing the occurrence of such processes in living neural systems \cite{Fremaux2010, Vasilaki2009}.

Beyond neuroscience, three-factor learning is explored in domains of machine learning, robotics, and neuromorphic computing \cite{Richards2019, Bellec2019, Potjans2009, Mozafari2019_2}. Traditional Hebbian and STDP-based learning often struggle with credit assignment over long timescales and stability, whereas three-factor learning integrates a modulatory signal that refines synaptic weight updates based on task-relevant feedback \cite{Kim2012, Usher2002}. These algorithms are particularly promising for tasks that require real-time decision-making, continuous learning, and resilience to environmental changes.

Robotics and cognitive modeling have also benefited from three-factor learning. Neuromodulated SNNs enable adaptive motor control, navigation, and sensor fusion, allowing agents to operate effectively in dynamic environments \cite{Parussel2007, Schmidgall2023_2, Alnajjar2009, Schmidgall2023}. Many studies develop neuromorphic controllers that incorporate reward-modulated plasticity for reinforcement learning, optimizing behavior through experience-dependent synaptic changes.

In computer vision and sensory processing with SNNs, three-factor learning has been applied to pattern recognition, object classification, and feature extraction \cite{Mozafari2019, Liu2021}. Compared to pure STDP, these approaches improve generalization and robustness, particularly in unsupervised or reinforcement learning settings. Other research explores affective computing, where neuromodulation is used to simulate adaptive emotional responses in artificial intelligence systems, influencing decision-making and learning strategies \cite{Talanov2015, Parussel2007}.

Lastly, a growing area of interest is neuromorphic hardware, where SNNs with three-factor learning are being implemented on specialized architectures for energy-efficient computation \cite{Potjans2009, Mozafari2019, Frenkel2019Odin, Frenkel2022Reckon}. Such solutions enable effective inference and on-chip learning, crucial in domains such as robotics.
In the following sections, we discuss the topics related to dedicated hardware for three-factor learning.

\begin{figure}[h]
    \centering
    \includegraphics[width=0.8\textwidth]{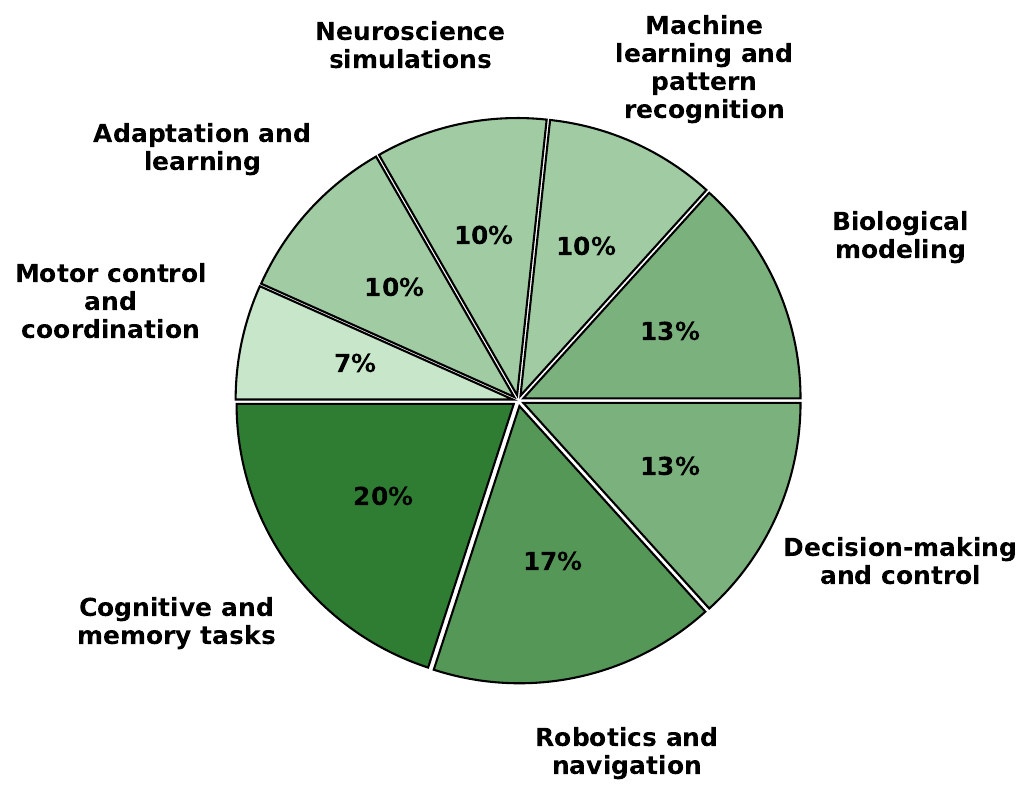}
    \caption{Application domains of SNN research reviewed in this work.}
    \label{fig:general_tasks}
\end{figure}

\begin{figure}[ht]
    \centering
    \includegraphics[width=0.8\textwidth]{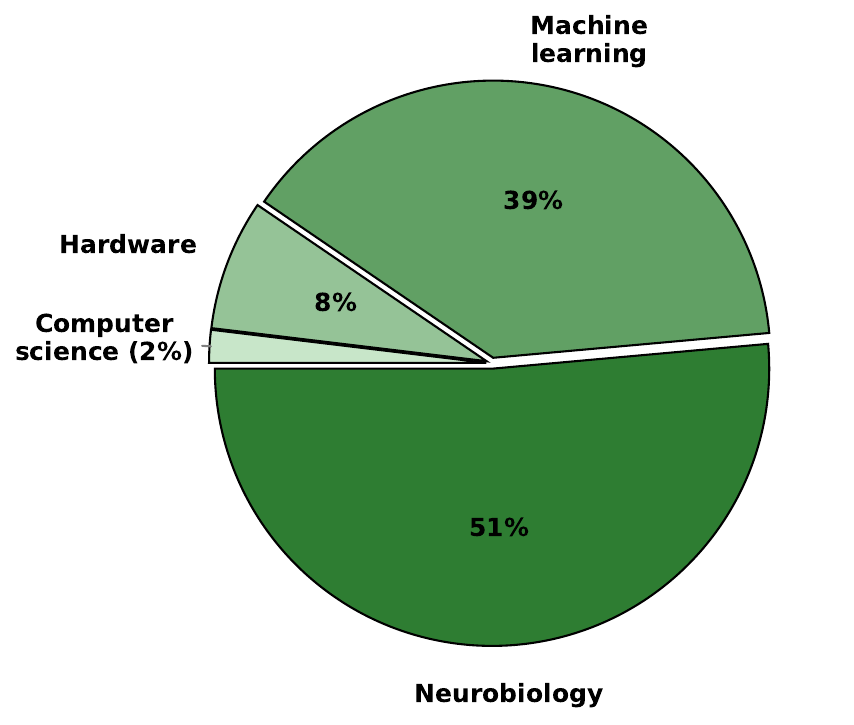}
    \caption{Primary scientific domain across reviewed research articles.}
    \label{fig:primary-domains}
\end{figure}

In Fig. \ref{fig:general_tasks}, we try to summarize the distribution of the research domains in the papers we focus on in this review. Additionally, in Fig. \ref{fig:primary-domains}, we show the distribution of scientific disciplines that are predominant across the revieved papers. In the context of the theoretical domain division in SNN research,  presented in Fig. \ref{fig:4-pie-plot}, we can see that the field of three factor learning is predominantly analyzed from the perspective of neurobiology and machine learning, leaving the electronics (hardware) and computer science (computational aspect) relatively underrepresented, showing the need for further research.

\subsection{Biological Insights}

Advances in neuroscience have contributed to our understanding of neuromodulation, synaptic plasticity, and neural circuit function. From the perspective of three-factor learning, knowledge on how neuromodulators such as dopamine, acetylcholine, norepinephrine, and serotonin influence neural activity, synaptic plasticity, and learning \cite{Marder2002, Pedrosa2017} has greatly influenced the development of such algorithms. In biological systems, neuromodulators influence plastic changes based on reward signals, errors, and contextual information \cite{Gerstner2018_2, Kuśmiesz2017}. Studies have shown that neuromodulators can alter the shape and polarity of STDP windows and regulate neuron excitability, firing patterns, and tuning curves \cite{Foncelle2018, Pawlak2010}. Specific findings include the roles of dopamine in reward processing, motivation, and memory robustness, as well as acetylcholine's contribution to attention and learning rate \cite{Avery2014, Lavigne2008}. Furthermore, research has highlighted the interaction of multiple neuromodulators, emphasizing their role in coordinating different aspects of cognitive functions and behavior \cite{Fremaux2016, Belkaid2020}.

Synaptic plasticity research has further deepened our understanding of mechanisms such as STDP and its variations, such as reward-modulated STDP (R-STDP) \cite{Mozafari2019, Zenke2015}. Neuromodulators have been shown to gate or modulate STDP, influencing synaptic changes by regulating calcium influx and excitatory-inhibitory balance in neural circuits \cite{Cutsuridis2011, Aljadeff2019}. Studies have also explored how different brain regions, such as the hippocampus, cortex, and basal ganglia, contribute to cognitive functions such as working memory and attention \cite{Szatm2010, Tiesinga2001}. Models have demonstrated how cholinergic and GABAergic modulation enhance visual attention and memory stability \cite{Avery2012, Avery2014}.

Neuromodulation has been shown to play an important role in homeostatic plasticity, the crucial set of mechanisms that maintain stable neural network function in the face of ongoing synaptic plasticity and activity changes \cite{turrigiano1999homeo, Tien2018homeoDev}. While synaptic (Hebbian) plasticity is essential for learning, homeostatic plasticity counteracts potential instability by regulating neuronal excitability and synaptic strength \cite{turrigiano1999homeo, zenke2017homeo}. This regulation prevents runaway potentiation or depression, ensuring that the activity of the system remains within a functional operating range \cite{Fox2017Stability, menesse2022criticality}. Key homeostatic mechanisms include synaptic scaling, which globally adjusts synaptic strength, and intrinsic plasticity, which modifies the intrinsic excitability of a neuron \cite{turrigiano2008tuningHomeo, Zhang2003homeo}. Neuromodulators influence these homeostatic processes. For instance, dopamine can modulate synaptic scaling and intrinsic excitability, affecting the stability and plasticity of developing neural circuits \cite{Strehl2018homeoDopamine, Morozova2016intrinsic}. Serotonin plays a role in the regulation of excitation-inhibition balance, a critical aspect of network homeostasis \cite{Tien2018homeoDev}. Acetylcholine contributes to firing rate homeostasis and interacts with synaptic scaling mechanisms \cite{Picciotto2012acetylcholine, bottorff2024firingHomeo}.
Although the primary form of plasticity discussed in machine learning with SNNs and in this article is related to synaptic plasticity, we note the importance of closer evaluation of homeostatic plasticity when developing novel algorithms.

The interplay between neuromodulators, synaptic plasticity, and homeostatic mechanisms underscores the complexity of biological learning. Three-factor learning algorithms in SNNs, inspired by these principles, offer a powerful framework to capture this complexity. However, to truly emulate biological intelligence, future research must move beyond isolated mechanisms and strive for a more holistic integration. This includes developing models that account for dynamic interactions between different neuromodulators, context-dependent modulation of STDP, and the stabilizing role of homeostatic processes.

\subsection{Applications in AI and Robotics}

Neuromodulation and three-factor learning have influenced advances in AI and robotics, with applications focusing on adaptive control and navigation. Studies highlight how neuromodulated learning enables robots to navigate, avoid obstacles, and manipulate objects in dynamic environments \cite{Sporns2002, Alnajjar2009}. Adaptive robotic control can be achieved through three-factor learning rules in SNNs, allowing robots to learn and adjust to new terrains and tasks in real time \cite{Schmidgall2023_2, Barry2022}. Some models use hierarchical control structures inspired by biological systems, such as the nervous system of \textit{Aplysia}, to enhance autonomous navigation \cite{Florian2005, Nolan2010}. In addition, emotion-modulated reinforcement learning has been explored to improve robot adaptability by adjusting learning rates and reward predictions based on neuromodulatory influences such as dopamine and acetylcholine \cite{Talanov2015, Belkaid2020}. The integration of real-world sensors with neural networks further supports adaptive behavior in complex environments, where rapid adaptability and online learning is crucial\cite{Kopsick2022, Sutton2021}

Neuromodulation in SNNs can also be achieved using reinforcement learning (RL), an established method in the field of robotics and autonomous systems \cite{Kober2013RLSurvey}. Actor-critic frameworks and R-STDP can be used to improve temporal credit assignment and decision-making processes \cite{Durstewitz, Usher2002}. Continuous-time RL mechanisms, combined with working memory features, enable evidence accumulation and better control of agents in dynamic scenarios \cite{Zambrano2021, Liu2021}. Cognitive and affective AI applications focus on neuromodulated architectures that simulate emotional influences, using neurotransmitter analogs such as dopamine and serotonin to improve creativity, decision making, and memory allocation \cite{Parussel2007, Talanov2015}. Lifelong learning is also supported by mechanisms such as surprise-modulated plasticity and controlled forgetting through dopaminergic modulation \cite{Brzosko2019, Allred2020}. Additional studies apply these innovations to pattern recognition, image classification, and decision-making tasks, often optimizing neuromorphic hardware models to improve energy efficiency and scalability \cite{Mozafari2019, Vigneron2020}. These advances reflect a multidisciplinary effort to create biologically inspired, robust, and adaptive systems capable of real-time learning and adaptation in uncertain environments. Thus, SNNs using three-factor learning show promise for advancing the domain of edge devices and robotics, especially because of their remarkable energy efficiency and adaptability.




\subsection{Scalability Considerations}

Scalability is a critical challenge in all computational methods, including SNNs. However, in the domain of SNNs, measurement of computational complexity and required resources is much more challenging than for any software deployed on CPUs or GPUs. The reason for this is the unique and asynchronous mode of operation of these networks, as neuronal signals are propagated sparsely over time. Thus, full exploitation of their properties is tightly coupled with the neuromorphic hardware that is used to deploy them. The co-design of software and hardware in SNNs is beyond the scope of this review, yet the awareness of its importance is growing \cite{Barrows2025Codesign, Oltra2021CoDesign, Frenkel2022Reckon, Frenkel2019Odin}.

Currently, most research in the domain of SNNs and three-factor learning either omits the computational complexity of proposed algorithms or tries to summarize it using the theoretical number of operations performed during run-time. However, standard complexity measures, such as floating point operations (FLOPs) commonly used in deep learning, are insufficient for SNNs due to their fundamentally different mode of computation. Unlike ANNs, which perform dense matrix multiplications at each layer, SNNs operate in an event-driven manner, where computations are sparse and depend on spike occurrences.

A common alternative is to count the number of accumulate operations, which refer to additions performed when integrating spikes incoming to a neuron \cite{chen2023trainingspikeneuralnetworks}. This contrasts with traditional ANNs, where operations typically involve multiply-accumulate (MAC) computations due to weight multiplications in dense layers. In some cases, authors rely also on classical big-O complexity analysis \cite{Quintana2024ETLP}.

Although useful for rough estimations, such methods remain limited because they do not account for hardware-specific optimizations, memory constraints, or differences in execution models, all of which can significantly impact the real-world efficiency of SNN implementations \cite{Frenkel2022Reckon, Orchad2021Loihii2, Rubino2023AnalogChip}.

Scalability and computational requirements are necessary to fully evaluate the system's usefulness when deployed, therefore establishing reliable metrics is necessary. We envision that in the future, measuring computational efficiency and scalability of SNNs together with used learning algorithms will inherently depend on the used neuromorphic platform. 



\subsection{Encoding Methods}
Methods of encoding analyzed data in spike trains that serve as input to SNN play a fundamental role in determining their performance and efficiency \cite{Guo2021CodingSNNS}. Neuroscience research has discovered that neurons use numerous encoding schemes in the brain \cite{SHIMAZAKI2025Coding, Rolls2011Encoding,Gerstner1997Encoding,Cariani2022Time}. While important, their full description is beyond the scope of this review. Thus, we will briefly describe only selected ones to highlight their trade-offs and complementarity, as well as popularity in three-factor learning research.

\textit{Rate encoding} is the most widely used encoding method in SNN research, in general. Represents information through the frequency of spikes, making it simple and compatible with neuromorphic hardware. It is also easily used for converting non-temporal data into the spiking representation. Using the example of static images, the intensities of individual pixels are treated as probabilities of spike occurrence in a given timestep.

\textit{Temporal encoding} methods leverage precise spike timing to convey information. The exist in many variations, but the core idea behind them is to emphasize the spikes that arrive earlier as the ones carrying more information \cite{thorpe1990spike}. Temporal encoding methods usually lead to lower computational complexity as the network emits a lower number of spikes \cite{Guo2021CodingSNNS}.


Lastly, we note the idea of \textit{fully adaptive encoding}. It refers to the set of methods that employ parametrized neural network layers that can learn the spike representation of input data \cite{Schmidgall2023}. It is still rarely used among SNN researchers, yet neuroscientific evidence shows its importance in biological neural systems, indicating the possibility of further exploration \cite{Fairhall2001Adaptivecoding}.

It is important to note that the type of encoding can be related to input data encoding or intraneuron communication in the network. However, most often the same encoding is applied for both cases.

\begin{figure}[h]
    \centering
    \includegraphics[width=0.8\textwidth]{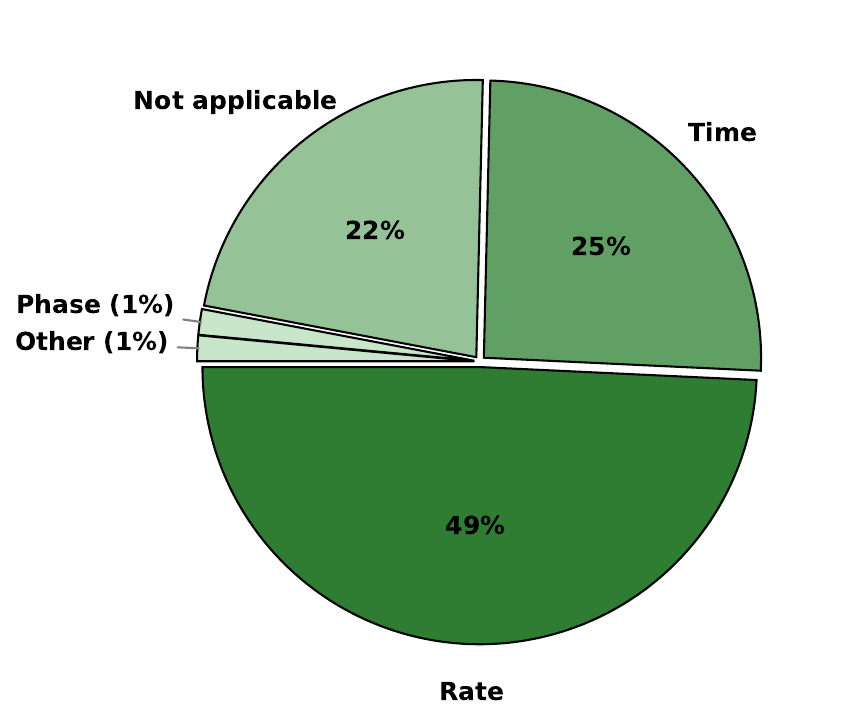}
    \caption{Overview of input encoding methods in SNNs, highlighting the predominance of rate encoding and the increasing adoption of time-based and hybrid encoding strategies.}
    \label{fig:encoding}
\end{figure}

Each encoding strategy presents trade-offs between efficiency, biological plausibility, and ease of hardware implementation. Figure~\ref{fig:encoding} provides a summary of the encoding methods used in the literature analyzed in this review. We consider only the type of input encoding which in most cases also translates to neuronal communication in the network.


The distribution of encoding methods in Figure~\ref{fig:encoding} underscores the widespread reliance on \textit{rate encoding}, which constitutes almost \textbf{ 50\%} of the approaches used in the research of the analyzed papers. This prevalence comes from its straightforward implementation and compatibility with neuromorphic hardware, despite its relatively lower temporal precision and increased computational cost \cite{Guo2021CodingSNNS}. \textit{Time-based encoding} follows as the second most utilized method in \textbf{25. 4\%}, reflecting the increasing emphasis on spike timing as a means of improving computational efficiency. 
\textit{Phase encoding} and \textit{adaptive encoding} was used in only \textbf{1.5\%} articles for both methods. Although beneficial, their lower popularity may indicate that the use of such encoding methods is yet to be explored.
The remaining \textbf{22.4\%} of articles were related to experimental neuroscience, simulations of neuronal dynamics, or other studies where input encoding was either not explicitly stated or not directly relevant.


This distribution suggests that, while rate-based encoding remains dominant, alternative strategies, particularly time-based approaches, are gaining traction as researchers explore more efficient and biologically plausible representations of neural information. This is especially relevant when considering the deployment on specialized hardware, where relying solely on rate coding can lead to increased computational cost \cite{Guo2021CodingSNNS}. Additionally, increasing the expressiveness of SNNs would require further exploration for determining optimal neural coding patterns both for input data and neuronal communication.

Finally, we emphasize that uniqueness of three-factor learning methods lies also in their general applicability for synaptic plasticity rules irrespective of chosen encoding method. 

\subsection{Hardware and Computational Platforms}

The development in the design and manufacturing of neuromorphic hardware has led to the emergence of numerous applications that deploy SNNs on such chips \cite{Ivanov2022Neuromorphic}. This choice of computing platforms significantly impacts the scalability and efficiency of SNN simulations. While traditional platforms like CPUs and GPUs dominate, neuromorphic hardware is gaining attention for its potential in energy-efficient processing.

Despite the growing number of SNN applications on neurmorphic chips, examples of application of three-factor learning on such chips are limited. In a work by Mikaitis et al. \cite{Mikaitis2018threeFactorSpinnaker}, where the authors show the effectiveness of this learning rule in solving the problem of credit assignment in the Pavlovian conditioning task on the Spinnaker \cite{Furber2014Spinnaker} chip. The proposed solution was compared with the GPU-based alternative, where neuromorphic implementation has shown a reduced runtime when scaling the number of synapses. Rostami et al. \cite{Rostami2022EpropSpinnaker} show an implementation of eprop in the Spinnaker2 prototype \cite{gonzalez2024spinnaker2}. They compare the three-factor method with backpropagation through time (BPTT, matching the performances of ANN models on the Google Speech Commands dataset \cite{Warden2018Commands}. Uludağ et al. \cite{Uludag2024Loihi2basal} used Loihi2 \cite{Orchard2021Loihii2} to create a model inspired by the basal ganglia to solve the Go / No-Go task, where synaptic plasticity was modulated by a signal that mimics the role of dopamine. 

Growing body of work showcases the effectiveness of three-factor learning on custom-made neuromorphic platforms.
Recently, Frenkel and Indiveri introduced the ReckOn neuromorphic accelerator to train recurrent neural networks \cite{Frenkel2022Reckon}. This chip also enables three-factor learning based on the adapted eprop algorithm. They demonstrate the feasibility of on-chip training via the aforementioned algorithm to obtain a network solving navigation task similar to effective as with BPTT. In previous work introducing ODIN and SPOON chips \cite{Frenkel2020A2C,Frenkel2019Odin}, Frenkel et al. also demonstrated a successful deployment of reward-based on chip learning to perform digit classification on the MNIST dataset \cite{Deng2012mnist}. Other research groups have also proposed their own chips with three-factor online learning capabilities, showing their effectiveness and energy efficiency in the MNIST classification \cite{Cen2019chip,Park2019chip,Buhler2017chip}.

Relative scarcity of solutions implementing thee-factor learning on neuromorphic chips points towards a possible unexplored research direction. Most modern neuromorphic systems support reward signals by design \cite{Li2024chipsSota}. Furthermore, a growing ecosystem of software development kits allows porting solutions based on three-factor learning to the dedicated hardware \cite{Lin2018loihiAPI}. 




\section{Limitations and Challenges}
Despite significant progress in three-factor learning for Spiking Neural Networks (SNNs), several limitations and challenges remain. These challenges span theoretical, computational, and practical aspects and affect the scalability, biological plausibility, and real-world applicability of current models. The primary concerns include the following.

\begin{itemize}
    \item \textbf{Simplified neuron models and network structures}: Many studies use simplified neuron models, such as the Hodgkin-Huxley model, leaky integrate-and-fire models, or Izhikevich models \cite{Tiesinga2001, Toporikova2011}. These models often lack the biological diversity and complexity of real neurons, including a limited diversity of receptor actions, simplified neuron morphologies, and a lack of detailed modeling of cellular processes \cite{Murphy-Royal2023, Foncelle2018}. Furthermore, network structures are often simplified, with limited spatial connectivity and inter-columnar connections, and sometimes consist of only a few layers \cite{Sporns2002, Zenke2015}. These simplifications can limit the generalizability of the findings and their applicability to real biological systems \cite{Gerstner2018_2, Sutton2021}.

    \item \textbf{Lack of real-world testing and global error propagation}: A significant number of studies rely on simulations and synthetic datasets \cite{Mozafari2019, Vigneron2020}, with a lack of real-world testing and empirical validation \cite{Liu2021, Richards2019}. Some models are tested in simple simulated environments and on simplified tasks, which limits their real-world applicability \cite{Alnajjar2009, Zambrano2021}. Furthermore, many models lack a clear mechanism for global error propagation, which is crucial for complex learning tasks \cite{Whittington2019, Richards2018}. Although some papers address credit assignment, most models do not fully implement it \cite{Richards2018, Fremaux2013}. Some models use simplified or indirect forms of supervision that may not be sufficient for complex tasks \cite{Kopsick2022, Najarro2020}.

    \item \textbf{Parameter tuning and scalability challenges}: Many models require careful parameter tuning for optimal performance \cite{Allred2020, Bellec2019}, and their performance can be sensitive to parameter choices \cite{Kim2012, Hoerzer2012}. Furthermore, many models have limited scalability and are not tested on large-scale networks or complex tasks \cite{Potjans2009, Florian2005}. Some models have high computational costs, which can limit their applicability \cite{Fremaux2010, Pedrosa2017}. Although some studies show scalability to some extent, they often highlight limitations when applied to more complex scenarios \cite{Szatm2010, Legenstein2009}. There is also a need for more efficient algorithms that can scale to larger and more complex networks \cite{Durstewitz, Pawlak2010}.
\end{itemize}

\section{Future Directions}  
Three-factor learning in SNNs presents exciting opportunities to bridge biological plausibility and machine learning efficiency. Future research should focus on optimizing neuromodulatory mechanisms for improved credit assignment, enhancing scalability for large networks, and integrating three-factor learning with modern deep learning frameworks. In addition, efforts should be made to validate these models with real-world data and neuromorphic hardware to enable practical applications in artificial intelligence, robotics, and cognitive computing. Cross-disciplinary collaborations will be essential in refining learning rules and expanding their applicability.

\subsection{Research Opportunities}  

Several sources highlight areas for improvement in the field of neuromodulation and plasticity in neural networks. A key challenge is to improve the scalability of current models. Many studies use simplified models and simulated data, and it is necessary to extend these models to larger, more complex networks and real-world datasets \cite{Zambrano2021, Bellec2019, Vigneron2020}. For example, while some models demonstrate scalability to a certain extent, they often note limitations when applied to more complex scenarios or real-world data \cite{Yi2023, Mozafari2019}.  

Another key research area involves the exploration of novel learning rules and architectures. Many studies introduce new learning methods or variations on existing ones, such as STDP, but these often require further validation and testing in diverse contexts \cite{Fremaux2016, Hoerzer2012, Yi2023}. For example, some studies propose new three-factor learning methods \cite{Gerstner2018, Kuśmiesz2017}, while others explore different ways to modulate STDP \cite{Foncelle2018, Pawlak2010}.  

There is also a need to better understand how neuromodulators can be used for credit assignment in deep networks \cite{Aljadeff2019, Brzosko2019, Whittington2019}. Some studies suggest that neuromodulators can act as a "third factor" in Hebbian learning, but the specific mechanisms and implementation details need further exploration \cite{Fremaux2013, Kuśmiesz2017, Richards2018}.  

Finally, the validation of computational models with experimental data is crucial. Many studies rely on simulations and lack direct empirical validation \cite{Edeline2012, Compte2003, Froemke2015}. Future research should focus on bridging the gap between theoretical models and experimental findings \cite{Pedrosa2017, Sutton2021, Schmidgall2023}.  

\subsection{Interdisciplinary Approaches}  

The sources strongly suggest that interdisciplinary collaboration is essential for progress in this field. The most successful studies often involve teams from diverse backgrounds, including neurobiology, machine learning, computer science, and robotics \cite{Vecoven2019, Zambrano2021, Schmidgall2023_2}. By combining expertise from different fields, researchers can gain a more comprehensive understanding of the complex interactions between neuromodulation, plasticity, and learning \cite{Richards2019, Vasilaki2009, Brzosko2019_2}.  

Specifically, integrating biological insights into artificial intelligence (AI) and machine learning (ML) models can lead to more robust and adaptable systems \cite{Mozafari2019, Yuan2019, Lavigne2008}. For example, modeling the effects of neuromodulators like dopamine, acetylcholine, and norepinephrine can lead to more sophisticated learning algorithms \cite{Gerstner2018, Talanov2015, Aljadeff2019}. The study of astrocytes and their role in neuromodulation also opens up new avenues for exploration \cite{Murphy-Royal2023}. Furthermore, understanding how the brain implements credit assignment, working memory, and decision-making processes can guide the development of novel AI architectures \cite{Gruber2003, Fremaux2013, Zambrano2021}.  

In summary, the future of this field lies in combining cutting-edge computational techniques with a deep understanding of biological mechanisms. By embracing interdisciplinary approaches, researchers can push the limits of what is possible and develop more powerful and biologically plausible AI systems \cite{Gerstner2018_2, Hoerzer2012, Sporns2002}.

\section{Conclusions}

This review has provided an overview of three-factor learning in Spiking Neural Networks (SNNs), highlighting its significance in bridging biological plausibility and machine learning efficiency. The inclusion of neuromodulatory signals as a third factor improves credit assignment, adaptive learning, and long-term synaptic modifications, making SNNs more suitable for real-world applications.

The key insights from this review emphasize the importance of interdisciplinary collaboration between neuroscience, artificial intelligence, and robotics. Advances in neuromorphic computing, biologically inspired algorithms, and novel encoding strategies continue to drive improvements in network scalability, learning efficiency, and cognitive modeling. Although significant progress has been made, challenges such as model validation with real-world data, scalability limitations, and computational efficiency remain critical research areas.

Future research should focus on integrating three-factor learning into scalable deep learning frameworks, optimizing neuromodulatory mechanisms for more biologically plausible credit assignment, and leveraging neuromorphic hardware for energy-efficient processing. By combining theoretical models with experimental validation and cross-domain collaboration, researchers can further refine learning rules and develop robust, adaptive systems.

Ultimately, the future of three-factor learning lies in its ability to integrate insights from biological systems into artificial intelligence, enabling more efficient, flexible, and human-like learning in neural networks. As the field advances, continued interdisciplinary efforts will be key to unlocking new possibilities in AI, cognitive computing, and robotics.

\section*{Acknowledgments}
This project has received funding from the European Union's Horizon 2020 research and innovation programme under grant agreement No 857533 and from the International Research Agendas Programme of the Foundation for Polish Science No MAB PLUS/2019/13. 

The publication was created within the project of the Minister of Science and Higher Education "Support for the activity of Centers of Excellence established in Poland under Horizon 2020" on the basis of the contract number MEiN/2023/DIR/3796. 
\bibliographystyle{plain}
\bibliography{references}

\begin{thebibliography}{100}

\bibitem{Aljadeff2019}
Johnatan Aljadeff, James D'amour, Rachel~E. Field, Robert~C. Froemke, and Claudia Clopath.
\newblock Cortical credit assignment by hebbian, neuromodulatory and inhibitory plasticity.
\newblock {\em arXiv preprint arXiv:1911.00307}, 2019.

\bibitem{Allred2020}
Jason~M. Allred and Kaushik Roy.
\newblock Controlled forgetting: Targeted stimulation and dopaminergic plasticity modulation for unsupervised lifelong learning in spiking neural networks.
\newblock {\em IEEE Transactions on Neural Networks and Learning Systems}, 31:1--14, 2020.

\bibitem{Alnajjar2009}
Fady Alnajjar, Indra~Bin Mohd~Zin, and Kazuyuki Murase.
\newblock A hierarchical autonomous robot controller for learning and memory: Adaptation in a dynamic environment.
\newblock In {\em Proceedings of the IEEE International Conference on Robotics and Biomimetics}, pages 1--6, 2009.

\bibitem{Avery2012}
Michael Avery, Jeffrey~L. Krichmar, and Nikil Dutt.
\newblock Spiking neuron model of basal forebrain enhancement of visual attention.
\newblock In {\em The 2012 International Joint Conference on Neural Networks (IJCNN)}, pages 1--8, 2012.

\bibitem{Avery2014}
Michael~C. Avery, Nikil Dutt, and Jeffrey~L. Krichmar.
\newblock Mechanisms underlying the basal forebrain enhancement of top‐down and bottom‐up attention.
\newblock {\em European Journal of Neuroscience}, 39(5):852--865, 2014.

\bibitem{Barrows2025Codesign}
Frank Barrows, Jonathan Lin, Francesco Caravelli, and Dante~R. Chialvo.
\newblock Uncontrolled learning: Codesign of neuromorphic hardware topology for neuromorphic algorithms.
\newblock {\em Advanced Intelligent Systems}, n/a(n/a):2400739, 2025.

\bibitem{Barry2022}
Martin Barry and Wulfram Gerstner.
\newblock Fast adaptation to rule switching using neuronal surprise.
\newblock {\em bioRxiv preprint}, 2022.

\bibitem{Beaumont2024checkpointing}
Olivier Beaumont, Lionel Eyraud-Dubois, Julien Herrmann, Alexis Joly, and Alena Shilova.
\newblock Optimal re-materialization strategies for heterogeneous chains: How to train deep neural networks with limited memory.
\newblock {\em ACM Trans. Math. Softw.}, 50(2), June 2024.

\bibitem{Belkaid2020}
Marwen Belkaid and Jeffrey~L. Krichmar.
\newblock Modeling uncertainty-seeking behavior mediated by cholinergic influence on dopamine.
\newblock {\em Neural Networks}, 125:10--18, 2020.

\bibitem{Bellec2019}
Guillaume Bellec, Franz Scherr, Elias Hajek, Darjan Salaj, Robert Legenstein, and Wolfgang Maass.
\newblock Biologically inspired alternatives to backpropagation through time for learning in recurrent neural nets.
\newblock {\em Neural Computation}, 31(2):417--443, 2019.

\bibitem{bottorff2024firingHomeo}
Juliet Bottorff, Sydney Padgett, and Gina~G. Turrigiano.
\newblock Basal forebrain cholinergic activity is necessary for upward firing rate homeostasis in the rodent visual cortex.
\newblock {\em Proceedings of the National Academy of Sciences}, 121(1):e2317987121, 2024.

\bibitem{Brzosko2019_2}
Zuzanna Brzosko, Susanna~B. Mierau, and Ole Paulsen.
\newblock A learning rule based on empirically-derived activity-dependent neuromodulation supports operant conditioning in a small network.
\newblock {\em Frontiers in Neural Circuits}, 13:Article 59, 2019.

\bibitem{Brzosko2019}
Zuzanna Brzosko, Susanna~B. Mierau, and Ole Paulsen.
\newblock Neuromodulation of spike-timing-dependent plasticity: Past, present, and future.
\newblock {\em Frontiers in Synaptic Neuroscience}, 11:25, 2019.

\bibitem{Buhler2017chip}
Fred~N. Buhler, Peter Brown, Jiabo Li, Thomas Chen, Zhengya Zhang, and Michael~P. Flynn.
\newblock A 3.43tops/w 48.9pj/pixel 50.1nj/classification 512 analog neuron sparse coding neural network with on-chip learning and classification in 40nm cmos.
\newblock In {\em 2017 Symposium on VLSI Circuits}, pages C30--C31, 2017.

\bibitem{Cariani2022Time}
Peter Cariani and Janet~M. Baker.
\newblock Time is of the essence: Neural codes, synchronies, oscillations, architectures.
\newblock {\em Frontiers in Computational Neuroscience}, 16, 2022.

\bibitem{Cen2019chip}
Gregory~K. Chen, Raghavan Kumar, H.~Ekin Sumbul, Phil~C. Knag, and Ram~K. Krishnamurthy.
\newblock A 4096-neuron 1m-synapse 3.8-pj/sop spiking neural network with on-chip stdp learning and sparse weights in 10-nm finfet cmos.
\newblock {\em IEEE Journal of Solid-State Circuits}, 54(4):992--1002, 2019.

\bibitem{chen2023trainingspikeneuralnetworks}
Guangyao Chen, Peixi Peng, Guoqi Li, and Yonghong Tian.
\newblock Training full spike neural networks via auxiliary accumulation pathway, 2023.

\bibitem{Compte2003}
Albert Compte, Maria~V. Sanchez-Vives, David~A. McCormick, and Xiao-Jing Wang.
\newblock Cellular and network mechanisms of slow oscillatory activity (<1 hz) and wave propagations in a cortical network model.
\newblock {\em Journal of Neurophysiology}, 89(5):2707--2725, May 2003.

\bibitem{Cramer2022SHD}
Benjamin Cramer, Yannik Stradmann, Johannes Schemmel, and Friedemann Zenke.
\newblock The heidelberg spiking data sets for the systematic evaluation of spiking neural networks.
\newblock {\em IEEE Transactions on Neural Networks and Learning Systems}, 33(7):2744--2757, 2022.

\bibitem{Cutsuridis2011}
Vassilis Cutsuridis.
\newblock Gaba inhibition modulates nmda-r mediated spike timing dependent plasticity (stdp) in a biophysical model.
\newblock {\em Neural Networks}, 24(7):646--654, 2011.

\bibitem{Deng2012mnist}
Li~Deng.
\newblock The mnist database of handwritten digit images for machine learning research [best of the web].
\newblock {\em IEEE Signal Processing Magazine}, 29(6):141--142, 2012.

\bibitem{Durstewitz}
D.~Durstewitz.
\newblock A few important points about dopamine's role in neural network dynamics.
\newblock {\em not specified in the paper}, n.d.

\bibitem{Edeline2012}
Jean-Marc Edeline et~al.
\newblock Beyond traditional approaches to understanding the functional role of neuromodulators in sensory cortices.
\newblock {\em Brain Research Reviews}, 66(1-2):1--9, 2012.

\bibitem{Fairhall2001Adaptivecoding}
Adrienne~L. Fairhall, Geoffrey~D. Lewen, William Bialek, and Robert~R. de~Ruyter~van Steveninck.
\newblock Efficiency and ambiguity in an adaptive neural code.
\newblock {\em Nature}, 412(6849):787--792, 2001.

\bibitem{Fink2011}
C.~Fink, V.~Booth, and M.~Zochowski.
\newblock Cellularly-driven differences in network synchronization propensity are differentially modulated by firing frequency.
\newblock {\em Frontiers in Computational Neuroscience}, 5:11, 2011.

\bibitem{Florian2005}
Răzvan~V. Florian.
\newblock A reinforcement learning algorithm for spiking neural networks.
\newblock {\em Neural Computation}, 17(10):2127--2141, 2005.

\bibitem{Foncelle2018}
Alexandre Foncelle, Alexandre Mendes, Joanna Jędrzejewska-Szmek, Silvana Valtcheva, Hugues Berry, Kim~T. Blackwell, and Laurent Venance.
\newblock Modulation of spike-timing dependent plasticity: towards the inclusion of a third factor in computational models.
\newblock {\em Frontiers in Synaptic Neuroscience}, 10:2, 2018.

\bibitem{Fox2017Stability}
Kevin Fox and Michael Stryker.
\newblock Integrating hebbian and homeostatic plasticity: introduction.
\newblock {\em Philosophical Transactions of the Royal Society B: Biological Sciences}, 372(1715):20160413, 2017.

\bibitem{Fremaux2016}
Nicolas Fr{\'e}maux and Wulfram Gerstner.
\newblock Neuromodulated spike-timing-dependent plasticity, and theory of three-factor learning rules.
\newblock {\em Frontiers in Computational Neuroscience}, 10:51, 2016.

\bibitem{Fremaux2010}
Nicolas Fr{\'e}maux, Henning Sprekeler, and Wulfram Gerstner.
\newblock Functional requirements for reward-modulated spike-timing-dependent plasticity.
\newblock {\em Neural Computation}, 22(2):192--210, 2010.

\bibitem{Fremaux2013}
Nicolas Fr{\'e}maux, Henning Sprekeler, and Wulfram Gerstner.
\newblock Reinforcement learning using a continuous time actor-critic framework with spiking neurons.
\newblock {\em Neural Computation}, 25(9):2255--2294, 2013.

\bibitem{Frenkel2022Reckon}
Charlotte Frenkel and Giacomo Indiveri.
\newblock Reckon: A 28nm sub-mm2 task-agnostic spiking recurrent neural network processor enabling on-chip learning over second-long timescales.
\newblock In {\em 2022 IEEE International Solid-State Circuits Conference (ISSCC)}, volume~65, pages 1--3, 2022.

\bibitem{Frenkel2019Odin}
Charlotte Frenkel, Martin Lefebvre, Jean-Didier Legat, and David Bol.
\newblock A 0.086-mm$^2$ 12.7-pj/sop 64k-synapse 256-neuron online-learning digital spiking neuromorphic processor in 28-nm cmos.
\newblock {\em IEEE Transactions on Biomedical Circuits and Systems}, 13(1):145--158, 2019.

\bibitem{Frenkel2020A2C}
Charlotte Frenkel, J.~D. Legat, and David Bol.
\newblock A 28-nm convolutional neuromorphic processor enabling online learning with spike-based retinas.
\newblock {\em 2020 IEEE International Symposium on Circuits and Systems (ISCAS)}, pages 1--5, 2020.

\bibitem{Froemke2015}
Robert~C. Froemke.
\newblock Plasticity of cortical excitatory-inhibitory balance.
\newblock {\em Annual Review of Neuroscience}, 38:195--219, 2015.

\bibitem{Furber2014Spinnaker}
Steve~B. Furber, Francesco Galluppi, Steve Temple, and Luis~A. Plana.
\newblock The spinnaker project.
\newblock {\em Proceedings of the IEEE}, 102(5):652--665, 2014.

\bibitem{Gerstner1997Encoding}
Wulfram Gerstner, Andreas~K. Kreiter, Henry Markram, and Andreas V.~M. Herz.
\newblock Neural codes: Firing rates and beyond.
\newblock {\em Proceedings of the National Academy of Sciences}, 94(24):12740--12741, 1997.

\bibitem{Gerstner2018_2}
Wulfram Gerstner, Marco Lehmann, Vasiliki Liakoni, Dane Corneil, and Johanni Brea.
\newblock Eligibility traces and plasticity on behavioral time scales: experimental support of neohebbian three-factor learning rules.
\newblock {\em Frontiers in Neural Circuits}, 12:88, 2018.

\bibitem{Gerstner2018}
Wulfram Gerstner, Marco Lehmann, Vasiliki Liakoni, Dane Corneil, and Johanni Brea.
\newblock Eligibility traces and plasticity on behavioral time scales: Experimental support of neohebbian three-factor learning rules.
\newblock {\em Frontiers in Neural Circuits}, 12:57, 2018.

\bibitem{gonzalez2024spinnaker2}
Hector~A. Gonzalez, Jiaxin Huang, Florian Kelber, Khaleelulla~Khan Nazeer, Tim Langer, Chen Liu, Matthias Lohrmann, Amirhossein Rostami, Mark Schöne, Bernhard Vogginger, Timo~C. Wunderlich, Yexin Yan, Mahmoud Akl, and Christian Mayr.
\newblock Spinnaker2: A large-scale neuromorphic system for event-based and asynchronous machine learning, 2024.

\bibitem{griffin2022caltech}
Gregory Griffin, Alex Holub, and Pietro Perona.
\newblock Caltech 256, kwi 2022.

\bibitem{Gruber2003}
Aaron~J. Gruber, Peter Dayan, Boris~S. Gutkin, and Sara~A. Solla.
\newblock Dopamine modulation in a basal ganglia-cortical network of working memory.
\newblock In {\em Advances in Neural Information Processing Systems}, volume~16, pages 935--942, 2003.

\bibitem{Guo2021CodingSNNS}
Wenzhe Guo, Mohammed~E. Fouda, Ahmed~M. Eltawil, and Khaled~Nabil Salama.
\newblock Neural coding in spiking neural networks: A comparative study for robust neuromorphic systems.
\newblock {\em Frontiers in Neuroscience}, 15, 2021.

\bibitem{Hasselmol2018}
M.E. Hasselmol and C.E. Stern.
\newblock A network model of behavioural performance in a rule learning task.
\newblock {\em not specified in the paper}, 2018.

\bibitem{Hebb1949}
Donald~O. Hebb.
\newblock {\em The Organization of Behavior: A Neuropsychological Theory}.
\newblock Wiley, New York, 1949.

\bibitem{Hoerzer2012}
Gregor~M. Hoerzer, Robert Legenstein, and Wolfgang Maass.
\newblock Emergence of complex computational structures from chaotic neural networks through reward-modulated hebbian learning.
\newblock {\em Neural Computation}, 24(8):2065--2094, 2012.

\bibitem{Ivanov2022Neuromorphic}
Dmitry Ivanov, Aleksandr Chezhegov, Mikhail Kiselev, Andrey Grunin, and Denis Larionov.
\newblock Neuromorphic artificial intelligence systems.
\newblock {\em Frontiers in Neuroscience}, 16, 2022.

\bibitem{Kim2012}
Dongbeom Kim, Yankang Chen, Pranit~S. Samarth, and Satish~S. Nair.
\newblock Computational study of the impact of neuromodulation on synaptic plasticity.
\newblock {\em Neural Computation}, 24(8):1952--1980, 2012.

\bibitem{Kitchenham2009Systematic}
Barbara Kitchenham, O.~{Pearl Brereton}, David Budgen, Mark Turner, John Bailey, and Stephen Linkman.
\newblock Systematic literature reviews in software engineering – a systematic literature review.
\newblock {\em Information and Software Technology}, 51(1):7--15, 2009.
\newblock Special Section - Most Cited Articles in 2002 and Regular Research Papers.

\bibitem{Kober2013RLSurvey}
Jens Kober, J.~Andrew Bagnell, and Jan Peters.
\newblock Reinforcement learning in robotics: A survey.
\newblock {\em The International Journal of Robotics Research}, 32(11):1238--1274, 2013.

\bibitem{Kopsick2022}
Jeffrey~D. Kopsick et~al.
\newblock Robust resting-state dynamics in a large-scale spiking neural network model of area ca3 in the mouse hippocampus.
\newblock In {\em Proceedings of the Cognitive Computation Conference}, pages 1190--1210, 2022.

\bibitem{Kuśmiesz2017}
L.~Kuśmiesz, T.~Isomura, and T.~Toyoizumi.
\newblock Learning with three factors: modulating hebbian plasticity with errors.
\newblock {\em PLoS Computational Biology}, 13(10):e1005682, 2017.

\bibitem{Lavigne2008}
Frédéric Lavigne and Nelly Darmon.
\newblock Dopaminergic neuromodulation of semantic priming in a cortical network model.
\newblock {\em Neuropsychologia}, 46(13):3074--3087, 2008.

\bibitem{Lecun2004norb}
Y.~LeCun, Fu~Jie Huang, and L.~Bottou.
\newblock Learning methods for generic object recognition with invariance to pose and lighting.
\newblock In {\em Proceedings of the 2004 IEEE Computer Society Conference on Computer Vision and Pattern Recognition, 2004. CVPR 2004.}, volume~2, pages II--104 Vol.2, 2004.

\bibitem{Legenstein2009}
Robert Legenstein, Steven~M. Chase, Andrew~B. Schwartz, and Wolfgang Maass.
\newblock Functional network reorganization in motor cortex can be explained by reward-modulated hebbian learning.
\newblock {\em Journal of Neuroscience}, 29(40):12733--12742, 2009.

\bibitem{leibe2003analyzing}
Bastian Leibe and Bernt Schiele.
\newblock Analyzing appearance and contour based methods for object categorization.
\newblock In {\em 2003 IEEE Computer Society Conference on Computer Vision and Pattern Recognition, 2003. Proceedings.}, volume~2, pages II--409. IEEE, 2003.

\bibitem{Li2024chipsSota}
Guoqi Li, Lei Deng, Huajin Tang, Gang Pan, Yonghong Tian, Kaushik Roy, and Wolfgang Maass.
\newblock Brain-inspired computing: A systematic survey and future trends.
\newblock {\em Proceedings of the IEEE}, 112(6):544--584, 2024.

\bibitem{Lillicrap2019bpttBrain}
Timothy~P Lillicrap and Adam Santoro.
\newblock Backpropagation through time and the brain.
\newblock {\em Current Opinion in Neurobiology}, 55:82--89, 2019.
\newblock Machine Learning, Big Data, and Neuroscience.

\bibitem{Lin2018loihiAPI}
Chit-Kwan Lin, Andreas Wild, Gautham~N. Chinya, Yongqiang Cao, Mike Davies, Daniel~M. Lavery, and Hong Wang.
\newblock Programming spiking neural networks on intel’s loihi.
\newblock {\em Computer}, 51(3):52--61, 2018.

\bibitem{Liu2021}
Yuhan~Helena Liu, Stephen Smith, Stefan Mihalas, Eric Shea-Brown, and Uygar Sümbül.
\newblock Cell-type–specific neuromodulation guides synaptic credit assignment in a spiking neural network.
\newblock {\em bioRxiv preprint}, 2021.

\bibitem{Lizbinski2018}
Kristyn~M. Lizbinski and Andrew~M. Dacks.
\newblock Intrinsic and extrinsic neuromodulation of olfactory processing.
\newblock {\em Frontiers in Cellular Neuroscience}, 11, 2018.

\bibitem{Marder2002}
Eve Marder and Vatsala Thirumalai.
\newblock Cellular, synaptic and network effects of neuromodulation.
\newblock {\em Neural Networks}, 15(4-6):479--493, 2002.

\bibitem{menesse2022criticality}
Gustavo Menesse, Bóris Marin, Mauricio Girardi-Schappo, and Osame Kinouchi.
\newblock Homeostatic criticality in neuronal networks.
\newblock {\em Chaos, Solitons \& Fractals}, 156:111877, 2022.

\bibitem{Mikaitis2018threeFactorSpinnaker}
Mantas Mikaitis, Garibaldi Pineda~García, James~C. Knight, and Steve~B. Furber.
\newblock Neuromodulated synaptic plasticity on the spinnaker neuromorphic system.
\newblock {\em Frontiers in Neuroscience}, 12, 2018.

\bibitem{Morozova2016intrinsic}
Ekaterina~O. Morozova, Denis Zakharov, Boris~S. Gutkin, Christopher~C. Lapish, and Alexey Kuznetsov.
\newblock Dopamine neurons change the type of excitability in response to stimuli.
\newblock {\em PLOS Computational Biology}, 12(12):1--36, 12 2016.

\bibitem{Mozafari2019}
Milad Mozafari, Mohammad Ganjtabesh, Abbas Nowzari-Dalini, Simon~J. Thorpe, and Timothée Masquelier.
\newblock Combining stdp and reward-modulated stdp in deep convolutional spiking neural networks for digit recognition.
\newblock {\em Frontiers in Neuroscience}, 13:1326, 2019.

\bibitem{Mozafari2019_2}
Milad Mozafari, Saeed~Reza Kheradpisheh, Timothée Masquelier, Abbas Nowzari-Dalini, and Mohammad Ganjtabesh.
\newblock First-spike-based visual categorization using reward-modulated stdp.
\newblock {\em IEEE Transactions on Neural Networks and Learning Systems}, 29(12):6178--6190, 2018.

\bibitem{Murphy-Royal2023}
Ciaran Murphy-Royal, ShiNung Ching, and Thomas Papouin.
\newblock A conceptual framework for astrocyte function.
\newblock {\em Nature Reviews Neuroscience}, 2023.

\bibitem{Najarro2020}
Elias Najarro and Sebastian Risi.
\newblock Meta-learning through hebbian plasticity in random networks.
\newblock {\em arXiv preprint}, arXiv:2003.03400, 2020.

\bibitem{Neftci2019Surrogate}
Emre~O. Neftci, Hesham Mostafa, and Friedemann Zenke.
\newblock Surrogate gradient learning in spiking neural networks: Bringing the power of gradient-based optimization to spiking neural networks.
\newblock {\em IEEE Signal Processing Magazine}, 36(6):51--63, 2019.

\bibitem{Nolan2010}
Christopher~R. Nolan, Gordon Wyeth, Michael Milford, and Janet Wiles.
\newblock The race to learn: spike timing and stdp can coordinate learning and recall in ca3.
\newblock {\em Hippocampus}, 21(6):647--660, 2010.

\bibitem{Oltra2021CoDesign}
Josep~Angel Oltra-Oltra, Jordi Madrenas, Mireya Zapata, Bernardo Vallejo, Diana Mata-Hernandez, and Shigeo Sato.
\newblock Hardware-software co-design for efficient and scalable real-time emulation of snns on the edge.
\newblock In {\em 2021 IEEE International Symposium on Circuits and Systems (ISCAS)}, pages 1--5, 2021.

\bibitem{Orchard2021Loihii2}
Garrick Orchard, E.~Paxon Frady, Daniel Ben~Dayan Rubin, Sophia Sanborn, Sumit~Bam Shrestha, Friedrich~T. Sommer, and Mike Davies.
\newblock Efficient neuromorphic signal processing with loihi 2.
\newblock In {\em 2021 IEEE Workshop on Signal Processing Systems (SiPS)}, pages 254--259, 2021.

\bibitem{Orchad2021Loihii2}
Garrick Orchard, Edward~Paxon Frady, Daniel Ben~Dayan Rubin, Sophia Sanborn, Sumit~Bam Shrestha, Friedrich~T. Sommer, and Mike Davies.
\newblock Efficient neuromorphic signal processing with loihi 2.
\newblock {\em CoRR}, abs/2111.03746, 2021.

\bibitem{Orchard2015Nmnist}
Garrick Orchard, Ajinkya Jayawant, Gregory~K. Cohen, and Nitish Thakor.
\newblock Converting static image datasets to spiking neuromorphic datasets using saccades.
\newblock {\em Frontiers in Neuroscience}, 9, 2015.

\bibitem{Page20220Prisma}
Matthew~J Page, Joanne~E McKenzie, Patrick~M Bossuyt, Isabelle Boutron, Tammy~C Hoffmann, Cynthia~D Mulrow, Larissa Shamseer, Jennifer~M Tetzlaff, Elie~A Akl, Sue~E Brennan, Roger Chou, Julie Glanville, Jeremy~M Grimshaw, Asbj{\o}rn Hr{\'o}bjartsson, Manoj~M Lalu, Tianjing Li, Elizabeth~W Loder, Evan Mayo-Wilson, Steve McDonald, Luke~A McGuinness, Lesley~A Stewart, James Thomas, Andrea~C Tricco, Vivian~A Welch, Penny Whiting, and David Moher.
\newblock The prisma 2020 statement: an updated guideline for reporting systematic reviews.
\newblock {\em BMJ}, 372, 2021.

\bibitem{Park2019chip}
Jeongwoo Park, Juyun Lee, and Dongsuk Jeon.
\newblock 7.6 a 65nm 236.5nj/classification neuromorphic processor with 7.5
\newblock In {\em 2019 IEEE International Solid-State Circuits Conference - (ISSCC)}, pages 140--142, 2019.

\bibitem{Parussel2007}
Karla Parussel and Lola Cañamero.
\newblock Biasing neural networks towards exploration or exploitation using neuromodulation.
\newblock {\em Adaptive Behavior}, 15(4):223--240, 2007.

\bibitem{Pascanu2013Vanishing}
Razvan Pascanu, Tomas Mikolov, and Yoshua Bengio.
\newblock On the difficulty of training recurrent neural networks.
\newblock In {\em Proceedings of the 30th International Conference on International Conference on Machine Learning - Volume 28}, ICML'13, page III–1310–III–1318. JMLR.org, 2013.

\bibitem{Pawlak2010}
V.~Pawlak, J.~Wickens, A.~Kirkwood, and J.~Kerr.
\newblock Timing is not everything: neuromodulation opens the stdp gate.
\newblock {\em Frontiers in Synaptic Neuroscience}, 2:146, 2010.

\bibitem{Pedrosa2017}
V.~Pedrosa and C.~Clopath.
\newblock The role of neuromodulators in cortical plasticity: A computational perspective.
\newblock {\em Current Opinion in Neurobiology}, 46:161--169, 2017.

\bibitem{Picciotto2012acetylcholine}
Marina~R. Picciotto, Michael~J. Higley, and Yann~S. Mineur.
\newblock Acetylcholine as a neuromodulator: Cholinergic signaling shapes nervous system function and behavior.
\newblock {\em Neuron}, 76(1):116--129, October 2012.

\bibitem{Potjans2009}
Wiebke Potjans, Markus Diesmann, and Abigail Morrison.
\newblock A spiking neural network model of an actor-critic learning agent.
\newblock In {\em Proceedings of the IEEE International Joint Conference on Neural Networks}, pages 2145--2152, 2009.

\bibitem{Quintana2024ETLP}
Fernando~M Quintana, Fernando Perez-Peña, Pedro~L Galindo, Emre~O Neftci, Elisabetta Chicca, and Lyes Khacef.
\newblock Etlp: event-based three-factor local plasticity for online learning with neuromorphic hardware.
\newblock {\em Neuromorphic Computing and Engineering}, 4(3):034006, aug 2024.

\bibitem{Richards2019}
Blake~A. Richards et~al.
\newblock A deep learning framework for neuroscience.
\newblock {\em Nature Neuroscience}, 22:1763--1774, 2019.

\bibitem{Richards2018}
Blake~A. Richards and Timothy~P. Lillicrap.
\newblock Dendritic solutions to the credit assignment problem.
\newblock {\em Current Opinion in Neurobiology}, 46:28--36, 2018.

\bibitem{Rolls2011Encoding}
Edmund~T. Rolls and Alessandro Treves.
\newblock The neuronal encoding of information in the brain.
\newblock {\em Progress in Neurobiology}, 95(3):448--490, 2011.

\bibitem{Rostami2022EpropSpinnaker}
Amirhossein Rostami, Bernhard Vogginger, Yexin Yan, and Christian~G. Mayr.
\newblock E-prop on spinnaker 2: Exploring online learning in spiking rnns on neuromorphic hardware.
\newblock {\em Frontiers in Neuroscience}, 16, 2022.

\bibitem{Rubino2023AnalogChip}
Arianna Rubino, Matteo Cartiglia, Melika Payvand, and Giacomo Indiveri.
\newblock Neuromorphic analog circuits for robust on-chip always-on learning in spiking neural networks.
\newblock In {\em 2023 IEEE 5th International Conference on Artificial Intelligence Circuits and Systems (AICAS)}, pages 1--5, 2023.

\bibitem{Rumelhart1986backprop}
David~E. Rumelhart, Geoffrey~E. Hinton, and Ronald~J. Williams.
\newblock Learning representations by back-propagating errors.
\newblock {\em Nature}, 323(6088):533--536, October 1986.

\bibitem{Schmidgall2023}
Samuel Schmidgall and Joe Hays.
\newblock Meta-spikepropamine: Learning to learn with synaptic plasticity in spiking neural networks.
\newblock {\em arXiv preprint}, arXiv:2305.12345, 2023.

\bibitem{Schmidgall2023_2}
Samuel Schmidgall and Joe Hays.
\newblock Synaptic motor adaptation: A three-factor learning rule for adaptive robotic control in spiking neural networks.
\newblock {\em arXiv preprint}, arXiv:2306.01906, 2023.

\bibitem{SHIMAZAKI2025Coding}
Hideaki Shimazaki.
\newblock Neural coding: Foundational concepts, statistical formulations, and recent advances.
\newblock {\em Neuroscience Research}, 2025.

\bibitem{Shrestha2019approximating}
Amar Shrestha, Haowen Fang, Qing Wu, and Qinru Qiu.
\newblock Approximating back-propagation for a biologically plausible local learning rule in spiking neural networks.
\newblock In {\em Proceedings of the International Conference on Neuromorphic Systems}, ICONS '19, New York, NY, USA, 2019. Association for Computing Machinery.

\bibitem{Soltani2006}
Alireza Soltani and Xiao-Jing Wang.
\newblock A biophysically based neural model of matching law behavior: melioration by stochastic synapses.
\newblock {\em Journal of Neuroscience}, 26(14):3731--3744, 2006.

\bibitem{Sporns2002}
Olaf Sporns and William~H. Alexander.
\newblock Neuromodulation and plasticity in an autonomous robot.
\newblock {\em Neural Networks}, 15(5-6):685--696, 2002.

\bibitem{Strehl2018homeoDopamine}
Andreas Strehl, Christos Galanis, Tijana Radic, Stephan~Wolfgang Schwarzacher, Thomas Deller, and Andreas Vlachos.
\newblock Dopamine modulates homeostatic excitatory synaptic plasticity of immature dentate granule cells in entorhino-hippocampal slice cultures.
\newblock {\em Frontiers in Molecular Neuroscience}, 11, 2018.

\bibitem{Sutton2021}
Nate~M. Sutton and Giorgio~A. Ascoli.
\newblock Spiking neural networks and hippocampal function: A web-accessible survey of simulations, modeling methods, and underlying theories.
\newblock {\em bioRxiv preprint}, 2021.

\bibitem{Suvrathan2018}
Aparna Suvrathan.
\newblock Beyond stdp—towards diverse and functionally relevant plasticity rules.
\newblock {\em Frontiers in Synaptic Neuroscience}, 10:45, 2018.

\bibitem{Szatm2010}
Botond Szatm{\'a}ry and Eugene~M. Izhikevich.
\newblock Spike-timing theory of working memory.
\newblock {\em Neural Computation}, 22(2):419--448, 2010.

\bibitem{Talanov2015}
M.~Talanov, J.~Vallverdú, S.~Distefano, M.~Mazzara, and R.~Delhibabu.
\newblock Neuromodulating cognitive architecture: Towards biomimetic emotional ai.
\newblock In {\em 2015 IEEE 29th International Conference on Advanced Information Networking and Applications (AINA)}, pages 587--592, 2015.

\bibitem{thorpe1990spike}
S.~J. Thorpe.
\newblock Spike arrival times: A highly efficient coding scheme for neural networks.
\newblock In R.~Eckmiller, G.~Hartmann, and G.~Hauske, editors, {\em Parallel Processing in Neural Systems and Computers}, pages 91--94. North-Holland, 1990.

\bibitem{Tien2018homeoDev}
Nai-Wen Tien and Daniel Kerschensteiner.
\newblock Homeostatic plasticity in neural development.
\newblock {\em Neural Development}, 13(1):9, June 2018.

\bibitem{Tiesinga2001}
P.~H.~E. Tiesinga, J.-M. Fellous, J.~V. Jos{\'e}, and T.~J. Sejnowski.
\newblock Optimal information transfer in synchronized neocortical neurons.
\newblock {\em Neurocomputing}, 38-40:397--402, 2001.

\bibitem{Toporikova2011}
Natalia Toporikova and Robert~J. Butera.
\newblock Two types of independent bursting mechanisms in inspiratory neurons: an integrative model.
\newblock {\em Journal of Computational Neuroscience}, 30(3):515--528, June 2011.

\bibitem{turrigiano1999homeo}
Gina~G Turrigiano.
\newblock Homeostatic plasticity in neuronal networks: the more things change, the more they stay the same.
\newblock {\em Trends in Neurosciences}, 22(5):221--227, 1999.

\bibitem{turrigiano2008tuningHomeo}
Gina~G. Turrigiano.
\newblock The self-tuning neuron: Synaptic scaling of excitatory synapses.
\newblock {\em Cell}, 135(3):422--435, October 2008.

\bibitem{Uludag2024Loihi2basal}
Recep~Buğra Uludağ, Serhat Çağdaş, Yavuz~Selim İşler, Neslihan~Serap Şengör, and İsmail Aktürk.
\newblock Bio-realistic neural network implementation on loihi 2 with izhikevich neurons.
\newblock {\em Neuromorphic Computing and Engineering}, 4(2):024013, jun 2024.

\bibitem{Usher2002}
Marius Usher and Eddy~J. Davelaar.
\newblock Neuromodulation of decision and response selection.
\newblock {\em Behavioral and Brain Sciences}, 25(1):1--19, 2002.

\bibitem{Vasilaki2009}
E.~Vasilaki, N.~Fr{\'e}maux, R.~Urbanczik, W.~Senn, and W.~Gerstner.
\newblock Spike-based reinforcement learning in continuous state and action space: when policy gradient methods fail.
\newblock {\em Neural Computation}, 21(4):999--1026, 2009.

\bibitem{Vecoven2019}
Nicolas Vecoven, Damien Ernst, Antoine Wehenkel, and Guillaume Drion.
\newblock Introducing neuromodulation in deep neural networks to learn adaptive behaviours.
\newblock {\em arXiv preprint}, cs.LG, 2018.

\bibitem{Vigneron2020}
Alex Vigneron and Jean Martinet.
\newblock A critical survey of stdp in spiking neural networks for pattern recognition.
\newblock {\em IEEE Access}, 8:12345--12356, 2020.

\bibitem{Warden2018Commands}
P.~{Warden}.
\newblock {Speech Commands: A Dataset for Limited-Vocabulary Speech Recognition}.
\newblock {\em ArXiv e-prints}, April 2018.

\bibitem{Werbos1990BPTT}
P.J. Werbos.
\newblock Backpropagation through time: what it does and how to do it.
\newblock {\em Proceedings of the IEEE}, 78(10):1550--1560, 1990.

\bibitem{Whittington2019}
J.~Whittington and R.~Bogacz.
\newblock Theories of error backpropagation in the brain.
\newblock {\em Trends in Cognitive Sciences}, 23(3):235--250, 2019.

\bibitem{Yi2023}
Zexiang Yi, Jing Lian, Qidong Liu, Hegui Zhu, Dong Liang, and Jizhao Liu.
\newblock Learning rules in spiking neural networks: A survey.
\newblock {\em Neurocomputing}, 531:163--179, 2023.

\bibitem{Yuan2019}
Mengwen Yuan, Xi~Wu, Huajin Tang, and Rui Yan.
\newblock Reinforcement learning in spiking neural networks with stochastic and deterministic synapses.
\newblock {\em Neural Networks}, 116:32--44, 2019.

\bibitem{Zambrano2021}
Davide Zambrano, Pieter~R. Roelfsema, and Sander Bohte.
\newblock Learning continuous-time working memory tasks with on-policy neural reinforcement learning.
\newblock {\em Neurocomputing}, 461:635--656, 2021.

\bibitem{Zenke2015}
Friedemann Zenke, Everton~J. Agnes, and Wulfram Gerstner.
\newblock Diverse synaptic plasticity mechanisms orchestrated to form and retrieve memories in spiking neural networks.
\newblock {\em Nature Communications}, 6:7922, 2015.

\bibitem{zenke2017homeo}
Friedemann Zenke, Wulfram Gerstner, and Surya Ganguli.
\newblock The temporal paradox of hebbian learning and homeostatic plasticity.
\newblock {\em Current Opinion in Neurobiology}, 43:166--176, 2017.
\newblock Neurobiology of Learning and Plasticity.

\bibitem{Zhang2003homeo}
Wei Zhang and David~J. Linden.
\newblock The other side of the engram: experience-driven changes in neuronal intrinsic excitability.
\newblock {\em Nature Reviews Neuroscience}, 4(11):885--900, November 2003.

\end{thebibliography}


\end{document}